%% file: acl2023.tex
\pdfoutput=1

\documentclass[11pt]{article}

\usepackage[backref=page]{hyperref}       

\usepackage{ACL2023}

\usepackage{times}
\usepackage{latexsym}

\usepackage[T1]{fontenc}

\usepackage[utf8]{inputenc}

\usepackage{microtype}

\usepackage{inconsolata}

\usepackage{bbm}
\usepackage[pdftex]{graphicx}
\usepackage{subcaption}
\usepackage{multirow}
\usepackage{amsthm,amsmath,amssymb}
\usepackage{mathtools}
\usepackage[inline]{enumitem}
\usepackage{tikz}
\usepackage{array}

\usepackage{booktabs}       
\usepackage{url}            

\usepackage{cleveref}

\input{math_commands}
\title{Revisiting the Architectures like Pointer Networks to Efficiently Improve the Next Word Distribution, Summarization Factuality, and Beyond}

\author{
Haw-Shiuan Chang\thanks{\, indicates equal contribution} \thanks{\, The work is done while the author was at UMass} \S  \, \,   Zonghai Yao\footnotemark[1] $\ddag$ \, \,   Alolika Gon$\ddag$ \, \, Hong Yu$\ddag$ \, \,  Andrew McCallum$\ddag$ \\ 
  $\ddag$ CICS, University of Massachusetts, Amherst\\
  \S Amazon Alexa AI \\
  \texttt{chawshiu@amazon.com, \{zonghaiyao,agon,hongyu,mccallum\}@cs.umass.edu}
}

\begin{document}
\maketitle
\begin{abstract}
Is the output softmax layer, which is adopted by most language models (LMs), always the best way to compute the next word probability? 
Given so many attention layers in a modern transformer-based LM, are the pointer networks redundant nowadays? In this study, we discover that the answers to both questions are no. This is because the softmax bottleneck sometimes prevents the LMs from predicting the desired distribution and the pointer networks can be used to break the bottleneck efficiently. Based on the finding, we propose several softmax alternatives by simplifying the pointer networks and accelerating the word-by-word rerankers. In GPT-2, our proposals are significantly better and more efficient than mixture of softmax, a state-of-the-art softmax alternative. In summarization experiments, without significantly decreasing its training/testing speed, our best method based on T5-Small improves factCC score by 2 points in CNN/DM and XSUM dataset, and improves MAUVE scores by 30\% in BookSum paragraph-level dataset.

\end{abstract}

\input{content/introduction.tex}

\input{content/background.tex}

\input{content/method_GPT2.tex}

\input{content/method_ED.tex}
\input{content/experiments.tex}

\input{content/analysis.tex}

\input{content/experiments_ED.tex}

\input{content/related_work.tex}
\section{Conclusion}
Since the transformer becomes the mainstream encoder and decoder for LMs, the output softmax layer seems to be the only reasonable option for computing the word probability distribution. Although being simple and efficient, the softmax layer is inherently limited while the existing solutions are relatively slow~\citep{chang2022softmax}. This work proposes a series of softmax alternatives that can improve the text generation models without increasing the computational costs significantly. Our experiments suggest that the main improvement of the pointer network on top of a transformer comes from breaking the softmax bottleneck. Our results also indicate that the alternatives could alleviate some problems of hallucination, repetition, and too generic generation. Furthermore, all of the proposed alternatives can be applied to the LMs that have already been pretrained using softmax without requiring retraining from scratch. For the practitioner, we recommend using all the partitioning methods together to get the best performance, or using only the simple context partition to keep the architecture simple while getting the majority of the gain. 

\section{Acknowledgement}
We thank Nadar Akoury and the anonymous reviewers for their constructive feedback.
This work was supported in part by the Center for Data Science and the Center for Intelligent Information Retrieval, 
in part by the Chan Zuckerberg Initiative under the project Scientific Knowledge Base Construction, 
in part by the IBM Research AI through the AI Horizons Network, 
in part using high performance computing equipment obtained under a grant from the Collaborative R\&D Fund managed by the Massachusetts Technology Collaborative, 
and in part by the National Science Foundation (NSF) grant numbers IIS-1922090 and IIS-1763618.
Any opinions, findings, conclusions, or recommendations expressed in this material are those of the authors and do not necessarily reflect those of the sponsor.


\input{content/ethics.tex}
\bibliography{custom}
\bibliographystyle{acl_natbib}

\clearpage

\appendix

\section{Appendix Overview}
\label{sec:appendix}
In the appendix, we first analyze our methods using more metrics in \Cref{sec:more_results} and describe what we learn from the results. Next, we provide some details of our methods and baselines in \Cref{sec:method_details}. Finally, we specify some experiment setups and hyperparameters in \Cref{sec:exp_details}.

\input{appendix/more_exp.tex}

\input{appendix/method_details.tex}

\input{appendix/exp_details.tex}

\end{document}

%% file: math_commands.tex
\usepackage{amsmath,amsfonts,bm}

\def\eqref#1{equation~\ref{#1}}

\def\1{\bm{1}}

\def\vb{{\bm{b}}}

\def\ve{{\bm{e}}}
\def\vf{{\bm{f}}}

\def\vh{{\bm{h}}}

\def\vq{{\bm{q}}}

\def\vv{{\bm{v}}}
\def\vw{{\bm{w}}}

\DeclareMathAlphabet{\mathsfit}{\encodingdefault}{\sfdefault}{m}{sl}
\SetMathAlphabet{\mathsfit}{bold}{\encodingdefault}{\sfdefault}{bx}{n}

%% file: content/introduction.tex
\section{Introduction}
When recurrent neural networks such as LSTM~\cite{hochreiter1997long} are the mainstream language model (LM) architecture, pointer networks, or so-called copy mechanisms~\citep{gu2016incorporating}, have been shown to improve the state-of-the-art LMs for next word prediction~\cite{MerityX0S17} and summarizations~\citep{cnn_dataset_split} by a large margin. However, after transformer~\citep{vaswani2017attention} becomes the dominating LM architectures, the pointer networks are rarely used in the state-of-the-art pretrained LMs.
One major reason is that the attention mechanism in every transformer layer can learn to copy the words from the context, so it seems to be redundant to add a copying mechanism on top of the transformer. 

\begin{figure}[t!]
\centering
\includegraphics[width=1\linewidth]{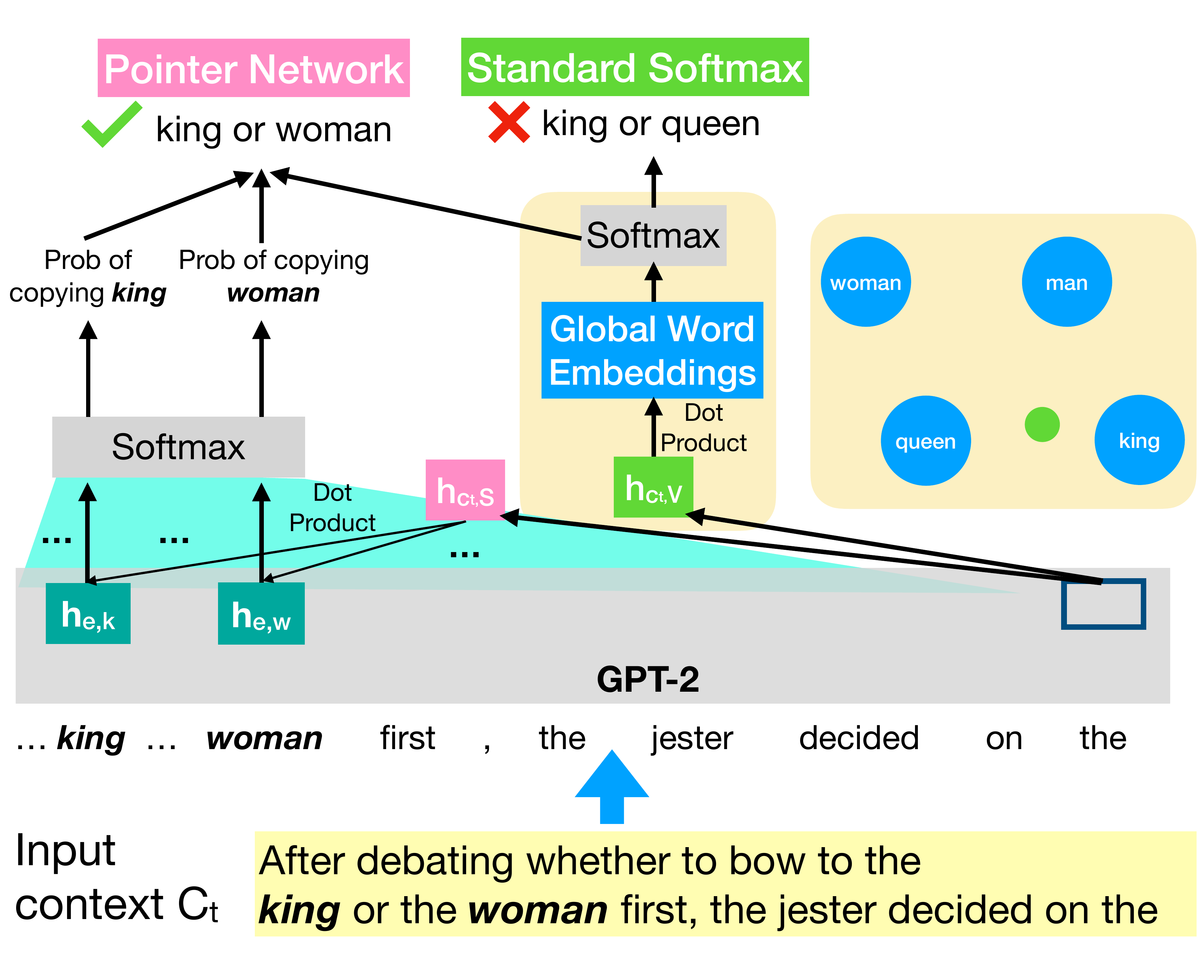}
\caption{Illustration of the softmax bottleneck and pointer network using an example from \citet{chang2022softmax}. GPT-2 cannot output both \textit{king} or \textit{woman} as the possible next word due to the parallelogram structure in the output word embedding space, while the pointer network could solve this by directly copying words from the context. The standard softmax estimate the probabilities of outputting \textit{king} and \textit{woman} by the dot products between the hidden state $\mathbf{h}_{c_t,V}$ and their global word embeddings. By contrast, The pointer networks compute the dot products between the projected current hidden state $\mathbf{h}_{c_t,S}$ and projected hidden states $\mathbf{h}_{e,.}$ for \textit{king} and \textit{woman} to estimate their probabilities.}
\label{fig:first_page}
\end{figure}

In this paper, we demonstrate that the architectures like pointer networks 
can still substantially improve the state-of-the-art transformer LM architectures such as GPT-2~\citep{radford2019language} and T5~\citep{2020t5} mainly due to breaking the bottleneck of their final softmax layer~\citep{yang2018breaking,chang2022softmax}. 

In \autoref{fig:first_page}, we illustrate a simple example from \citet{chang2022softmax} to explain the softmax bottleneck and why the pointer networks could alleviate the problem. When predicting the next word, most LMs would try to output a hidden state $\mathbf{h}_{c_t,V}$ that is close to all the next word possibilities. For example, when the next word should be either \textit{king} or \textit{woman} with similar probabilities, the ideal hidden state is supposed to be the average of the global output word embeddings of \textit{king} and \textit{woman}. However, there might be other interfering words (\textit{queen} and \textit{man} in this case) between the ideal next word candidates, which force the LM to output the wrong distribution.

To solve this problem, we can let the LMs predict the probability of copying the words in the context separately by paying attention to the previous hidden states~\cite{gu2016incorporating} and we call this kind of architecture pointer networks in this paper. That is, we can compute the dot products with the hidden states of \textit{king} $\mathbf{h}_{e,k}$ and the hidden states of \textit{woman} $\mathbf{h}_{e,w}$ rather than with their global output word embeddings in order to estimate the probabilities of copying these two words in the context. Our experiments show that the pointer networks consistently improve the performance of GPT-2 in next word prediction and the quality of summarization from T5 and BART. 

Contrary to the mainstream explanation in previous pointer network literature, we discover that most of the improvements in our experiments do not come from the attention mechanism. To study these improvements, we propose a very simple pointer network variant that does not use any previous hidden states and we show that the proposed method can achieve similar improvements. 

\begin{figure}[t!]
\centering
\includegraphics[width=1\linewidth]{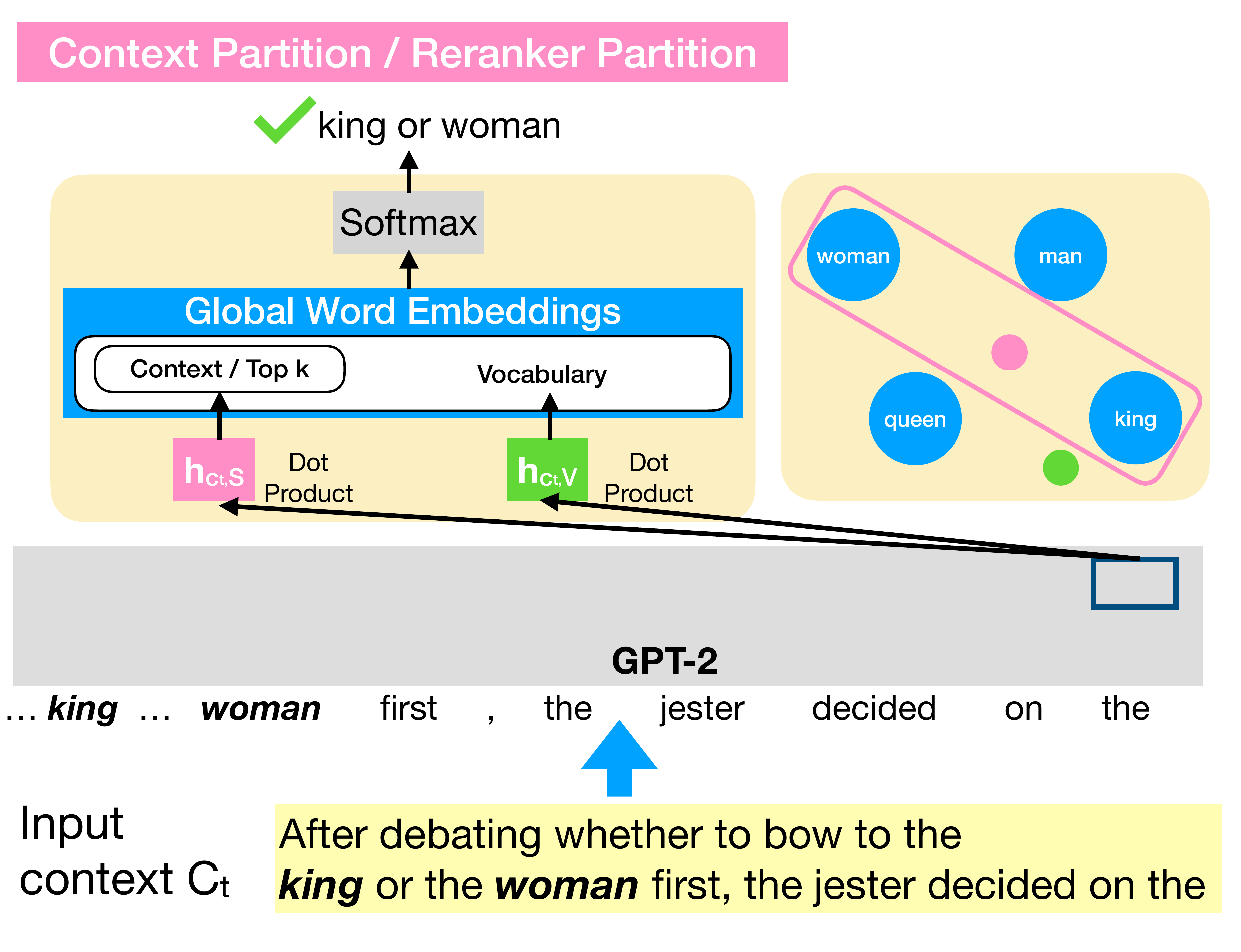}
\caption{We simplify the pointer network / reranker by using another embedding $\mathbf{h}_{c_t,S}$ for the words in the context / the top-k likely words. }
\label{fig:first_page2}
\end{figure}

As shown in \autoref{fig:first_page2}, we simply project the last hidden state into two embeddings. One embedding $\mathbf{h}_{c_t,S}$ is to compute the dot product with the context words, and $\mathbf{h}_{c_t,V}$ is for the dot product of the other words. Then, the GPT-2 can output the hidden state for context words $\mathbf{h}_{c_t,S}$ as the average embedding of the \textit{king} and \textit{woman} without interfered by the words of \textit{man} and \textit{queen} that are handled by $\mathbf{h}_{c_t,V}$. We call this method context partition. In addition to words in the context, we can also use another embedding for the top-k likely next words. This can be viewed as a very simple and efficient alternative to a reranker, so we call it reranker partition.

In our experiments, we show that the context partition performs similarly to pointer networks 
while 
combining a pointer network, context partition, and reranker partition would significantly outperform each individual method. Compared to the state-of-the-art solutions for alleviating the softmax bottleneck such as mixture of softmax~\citep{yang2018breaking,chang2022softmax}
, our proposed method is more efficient while achieving lower perplexity on GPT-2. Furthermore, we find that adding a very expensive word-by-word reranker only improves our method slightly, which suggested the difficulty of further improving the final softmax layer over the proposed alternatives.


In the text completion task using GPT-2, we find that the proposed softmax alternatives reduce hallucination by copying more proper nouns from the context even though we did not provide any part-of-speech information during training. In summarization, our methods and pointer networks output a more specific summary, increase the factuality, and consistently improve 9 metrics, especially 
in the smaller language models.
Finally, we show that the softmax bottleneck problem is not completely solved in GPT-3.5 in the limitation section.





\subsection{Main Contributions}
\begin{itemize}[leftmargin=.2in,topsep=0pt]
\setlength\itemsep{0.0em}
    \item We propose a series of efficient softmax alternatives that unify the ideas of pointer network, reranker, multiple embeddings, and vocabulary partitioning.\footnote{Our codes are released at \url{https://github.com/iesl/Softmax-CPR}}
    \item We evaluate the proposed softmax alternatives in text completion tasks and summarization tasks using various metrics to identify where our methods improve the most.
    \item Our experiments indicate 
    pointer networks and our proposed alternatives can still improve the modern transformer-based LMs. By breaking the softmax bottleneck, our methods learn to sometimes copy the context words to reduce generation hallucination and sometimes exclude the context words to reduce the repetition. Besides, we find that the softmax bottleneck problem won't be completely solved by the huge size of GPT-3.5.
\end{itemize}








%% file: content/background.tex
\section{Background}

Before introducing our method, we would first briefly review the problem we are solving and its state-of-the-art solutions.

\subsection{Softmax Bottleneck Problem}
\label{sec:bottleneck}

Most LMs use a softmax layer to compute the final probability of predicting the word $x$:
\begin{equation}
\label{eq:softmax}
P_{M}(x|c_t) = \frac{\text{exp}(\text{Logit}(x,c_t))}{\sum_{x'} \text{exp}(\text{Logit}(x',c_t))},
\end{equation}
where $c_t$ is the context words. Typically, the logit $\text{Logit}(x,c_t)=(\vh^M_{c_t})^T \vw_x$, $\vh^M_{c_t}$ is the $M$th-layer hidden state given the input context $c_t$ and $\vw_x$ is the output word embeddings for $x$.

One problem is that the output word embeddings $\vw_x$ are global and independent to the context. 
After pretraining, the similar words would have similar output word embeddings. 
However, the similarity structure in the word embedding space might prevent LMs from outputting the desired distribution. The parallelogram structure among the embeddings of \textit{king}, \textit{queen}, \textit{woman}, and \textit{man} is a simple example. \citet{chang2022softmax} generalize this observation and show that some words in a small subspace would create some multi-mode distributions that a LM cannot output using a single hidden state $\vh_{c_t}$ in the softmax layer.


\subsection{Mixture of Softmax Method}

To overcome the bottleneck, one natural solution is to have multiple hidden states and each hidden state corresponds to a group of possible words~\citep{yang2018breaking}. For example, we can have one hidden state for \textit{king} and another hidden state for \textit{woman}. 

One major concern of this mixture of softmax (MoS) approach is the computational overhead. MoS needs to compute the final softmax multiple times and merge their resulting distributions. That is, we need to compute the dot products between every hidden state and all the words in the vocabulary, which is expensive especially when the vocabulary size is large. 


\subsection{Multiple Input State Enhancement}
\label{sec:Mi}

In MoS, the multiple hidden states come from the linear projections of the last hidden state. \citet{chang2022softmax} point out that the total degree of freedom among the multiple hidden states is limited by the dimensionality of the hidden state.

To allow LMs to move multiple hidden states more freely, \citet{chang2022softmax} propose to concatenate the projection of a block of hidden state with the last hidden state $\vh_{c_t}^M$ so as to increase its dimensionality:
\begin{equation}
\label{eq:multi-hidden}
\vq_{c_t} = \vh_{c_t}^M \oplus  GELU\left(L^h(\oplus_{i,m} \vh_{c_{t-i}}^{M-m})\right),
\end{equation}
where $GELU$ is the non-linear transformation used in GPT-2 and $L^h$ is a linear transformation that allows us to consider more hidden states without significantly increasing the model size. $\oplus_{i,m} \vh_{c_{t-i}}^{M-m}$ is the concatenation of a block of hidden states. We set the block size to be 3x3 in our GPT-2 experiments and 1x3 in our summarization experiments (i.e., considering the last 3 hidden states in the last layer as shown in \autoref{fig:all_partition}).


%% file: content/method_GPT2.tex
\begin{table}[t]
\scalebox{0.62}{
\begin{tabular}{lc|ll}
                                  & Abbr.  & Partition ($S$) & Word Emb ($\ve_x$) \\ \hline
Context Partition                 &  C  & Decoder context & Global word emb  \\
Encoder Partition                 &  E & Encoder input & Global word emb  \\
PS (LD)~\citep{MerityX0S17} & \multirow{2}{*}{P} & Decoder context & Decoder state \\
PG (LE)~\citep{cnn_dataset_split} &   & Encoder input & Encoder state \\
Reranker Partition                & R & Top k & Global word emb  \\

\end{tabular}
}
\caption{Comparison of different softmax alternatives and their abbreviation (Abbr.) using \autoref{eq:logit_general}. PS: Pointer Sentinel. PG: Pointer Generator. LD: local decoder embedding. LE: local encoder embedding.} 
\label{tb:method_comparison}
\end{table}

\begin{figure*}[t!]
\centering
\includegraphics[width=1\linewidth]{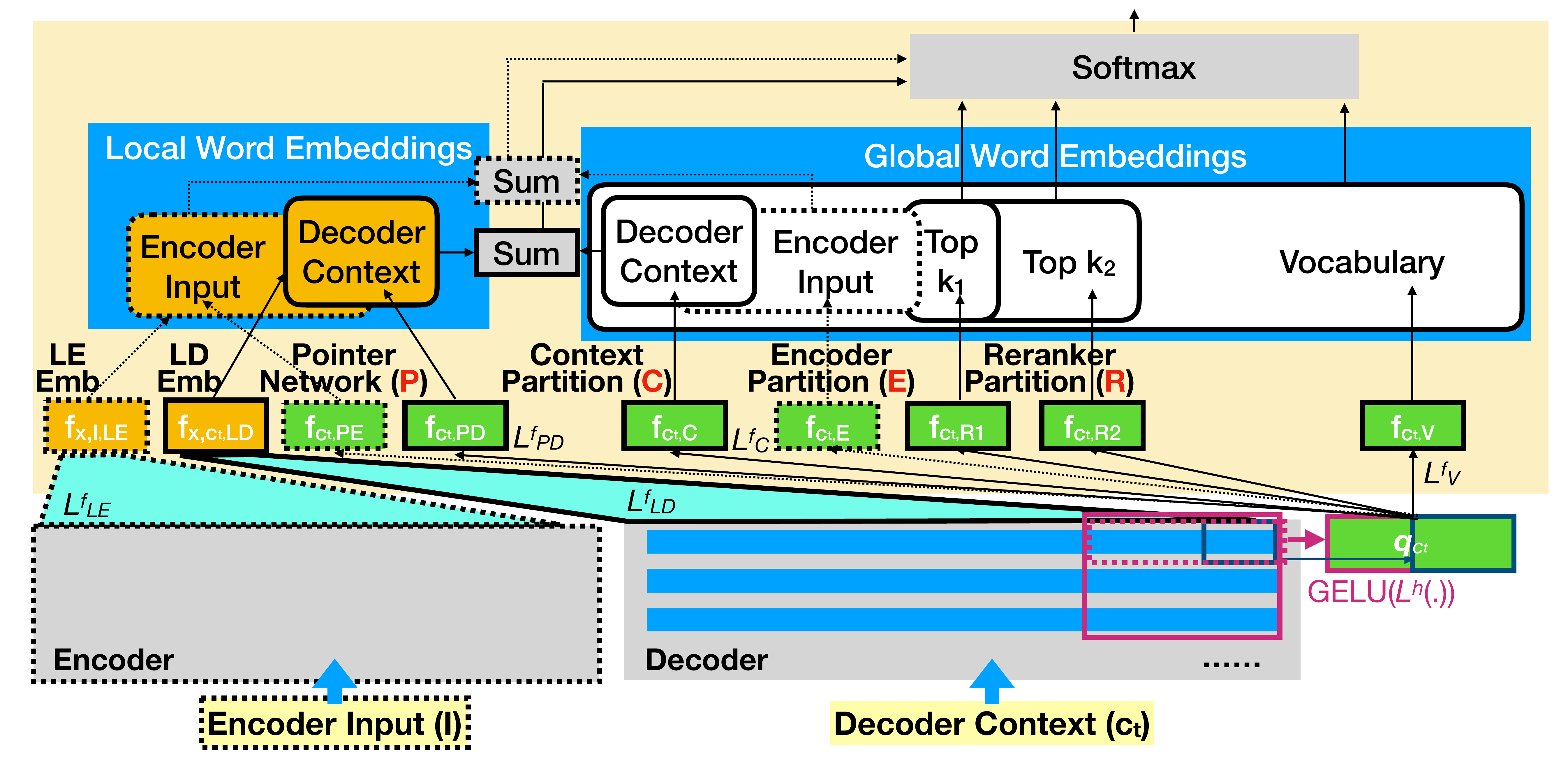}
\caption{
Architectures of our method for T5/BART that computes $\text{Logit}_{CEPR}$ in \autoref{eq:logit_CEPR}. In GPT-2, we use same architecture except that we take the 3x3 input hidden state block rather than the 1x3 block and there are no encoder-related components, which are marked by dotted lines.
}
\label{fig:all_partition}
\end{figure*}

\section{Methods}

To break the softmax bottleneck more efficiently compared to MoS, our overall strategy is simple. If we can identify a small partition of words that are very likely to become the next word, we can just compute the dot products between a hidden state and the embeddings of these likely words instead of all the words as in MoS. For example, if we can identify \textit{king} and \textit{woman} are much more likely to appear than \textit{queen} and \textit{man}, we can only compute the dot product between a hidden state and the embeddings of \textit{king} and \textit{woman} without being interfered by other words.

Specifically, when we compute the next word probability in \autoref{eq:softmax}, the logit of the word $x$ given the context $c_t$ 
\begin{equation}
\label{eq:logit_general}
\text{Logit}(x,c_t)=\left\{
\begin{matrix*}[l]
\vf_{c_t,S}^T \ve_x  \;\;\; \text{if} \; x \in S\\ 
\vf_{c_t,V}^T \vw_x \;\; \text{O/W}
\end{matrix*}\right.,
\end{equation}
where $\vf_{c_t,S}=L_S^f(\vq_{c_t})$ and $\vf_{c_t,V}=L_V^f(\vq_{c_t})$ are the linear projections of the hidden state concatenation $\vq_{c_t}$ in \autoref{eq:multi-hidden}. As shown in \autoref{tb:method_comparison}, different softmax alternatives have different ways of constructing this set $S$ and use different word embeddings $\ve_x$.


To simplify our explanation, we will focus on the decoder-only LM (i.e., GPT-2) first and extend our method to encoder-decoder LM (i.e., T5 and BART).

\subsection{GPT-2}
We will explain each softmax alternative individually and their connections to previous work such as pointer networks or rerankers. 



\subsubsection{Pointer Network (P) as Local Word Embedding}

Similar to Pointer Sentinel (PS)~\citep{MerityX0S17}, we treat the words in the context differently ($S=\{x| x \in c_t\}$) and let their word embeddings $\ve_x$ come from the previous hidden states:
\begin{equation}
\label{eq:local_emb}
\ve_{x} = \vf_{x,c_t,LD} = \frac{\sum_{i=1}^{t} \mathbbm{1}_{ {c_t^i}=x} L_{LD}^f(\vq_{c_t^i}) }{\sum_{i=1}^{t} \mathbbm{1}_{{c_t^i}=x}},
\end{equation}
where $c_t^i$ is the $i$th input words in the context $c_t$, $L_{LD}^f$ is a linear layer, and $\mathbbm{1}_{ {c_t^i}=x} = 1 \; \; \text{if} \; c_t^i=x$.

As a result, we can use the GPT-2 model to not only predict the hidden state $\vf_{c_t,S}=\vf_{c_t,PD}=L_{PD}^f(\vq_{c_t})$ and $\vf_{c_t,V}$ but also predict the word embedding of context words $\ve_{x}$. Unlike the global word embedding $\vw_x$, the local word embedding $\ve_{x}$ is context-dependent, so the LM can break the softmax bottleneck by adjusting the similarity of words based on the context. For example, GPT-2 could increase the similarity between $\ve_{\text{king}}$ and $\ve_{\text{woman}}$ to output the high probability for both words easily.

We call this version of pointer network local decoder (LD) embedding, which has some minor differences compared to PS~\citep{MerityX0S17} and other variants. For example, we merge their logits while PS merges their probabilities. PS does not do normalization when computing $\ve_{x}$. In our experiments, we would show that these pointer network variants all have very similar improvements in modern LMs.



\subsubsection{Context Partition (C)}
\label{sec:context_partition}

To understand the source of the improvements from pointer networks, we simplify their architectures by setting the word embedding $\ve_{x}=\vw_x$ and the partition $S$ is still the set of context words. Although much simpler, the LM with this context partition method can still break the softmax bottleneck by properly coordinating the hidden state specifically for the context words $\vf_{c_t,S}=\vf_{c_t,C}=L_{C}^f(\vq_{c_t})$ and the hidden state for other words $\vf_{c_t,V}$. Compared to the pointer network, one advantage of context partition is that the LM can still leverage the learned global word similarity when estimating the probabilities of context words.





\subsubsection{Reranker Partition (R)}
In some cases, the possible next words might not be mentioned in the context. For example, in the context \textit{My favorite actor is Ryan [MASK]}, the next word could be \textit{Reynolds}, \textit{Gosling}, or the last names of other \textit{Ryan}. Hence, using only the context partition does not completely solve the multimodal distribution problem.

Inspired by the idea of the reranker, we set $S$ to be the top $k$ words with the highest logits $\vf_{c_t,V}^T \vw_x$. In practice, finding an ideal $k$ could be difficult. When $k$ is small, the reranker partition might not include the very likely next word. When $k$ is large, the reranker partition might not be able to separate the output candidates and the interfering words. To alleviate the problem, we can have multiple reranker partitions and use different hidden state embeddings (e.g., $\vf_{c_t,R1}$ and $\vf_{c_t,R2}$) for different partitions.

\subsubsection{Hybrid Approach (CPR)}

Local embeddings in the pointer networks and global embeddings in the context partition are complementary. Using local embeddings is representational powerful while using global embedding can leverage the global similarity of words. Hence, we can combine the two methods by summing their dot products.

For the methods that use different $S$, we can simply determine an order of computing the dot products and let the later dot products overwrite the existing values. In our experiments, we always use the order illustrated in \autoref{fig:all_partition}. That is, we compute the logits ($\text{Logit}_{CPR}(x,c_t)$) by
\begin{equation}
\label{eq:logit_CPR}
\left\{
\begin{matrix*}[l]
\vf_{c_t,C}^T \vw_x + \vf_{c_t,PD}^T \vf_{x,c_t,LD} \;\; \text{if} \; x \in c_t\\ 
\vf_{c_t,R1}^T \vw_x \;\; \text{if} \; x \in W(k_1) - c_t \\
\vf_{c_t,R2}^T \vw_x \;\; \text{if} \; x \in W(k_2) - W(k_1) - c_t \\
\vf_{c_t,V}^T \vw_x \;\; \text{O/W}
\end{matrix*}\right.,
\end{equation}
where $W(k_2)$ is the top $k_2$ words with the highest $\vf_{c_t,V}^T \vw_x$ and $W(k_1)$ is the top $k_1$ words with the highest $\max(\vf_{c_t,V}^T \vw_x, \vf_{c_t,R2}^T \vw_x)$.






%% file: content/method_ED.tex
\subsection{T5 and BART}
In the encoder-decoder architectures, our local decoder embedding, context partition, and reranker partitions are still applicable. Besides, we can leverage the words in the encoder input to further improve the performance.

\subsubsection{Encoder Partition (E) and Local Encoder Embedding (P)}
Similar to the context partition, the encoder partition handles the words in the encoder input $I$ differently by setting $S=\{x | x \in I\}$ and using the global word embedding $\ve_x = \vw_x$. 

As in \autoref{eq:local_emb}, we can also let the hidden states in the last layer pass through another linear layer $L_{LE}^f()$ to
predict the embeddings of the words in the encoder input. The method is called local encoder (LE) embedding.


\subsubsection{Hybrid Approach (CEPR)}


Similar to GPT-2, we combine local encoder embedding and encoder partition for computing the probabilities of the words that are in the encoder context but not in the decoder context. As shown in \autoref{fig:all_partition}, we compute $\text{Logit}_{CEPR}(x,c_t)$ by
\begin{equation}
\label{eq:logit_CEPR}
\left\{
\begin{matrix*}[l]
\vf_{c_t,C}^T \vw_x + \vf_{c_t,PD}^T \vf_{x,c_t,LD} \;\; \text{if} \; x \in c_t\\ 
\vf_{c_t,E}^T \vw_x + \vf_{c_t,PE}^T \vf_{x,I,LE} \;\;\; \text{if} \;  x \in I - c_t\\ 
\vf_{c_t,R1}^T \vw_x \;\; \text{if} \; x \in W(k_1) - c_t - I \\
\vf_{c_t,V}^T \vw_x \;\;\; \text{O/W}
\end{matrix*}\right.,
\end{equation}
which is the same as \autoref{eq:logit_CPR} except that we add the encoder partition and local encoder embedding, and we remove the second reranker partition.

%% file: content/experiments.tex
\section{Experiments}

The pointer network was a popular technique in language modeling~\cite{MerityX0S17} and summarization~\citep{cnn_dataset_split}. Thus, we also focus on these two fundamental applications.

\begin{table*}[t!]
\centering
\scalebox{0.72}{
\begin{tabular}{c|cccc|cccc|}
&  \multicolumn{4}{c|}{GPT-2 Small} & \multicolumn{4}{c|}{GPT-2 Medium} \\
Model Name & Size & Time (ms)  & OWT ($\downarrow$) & Wiki ($\downarrow$)& Size & Time (ms) & OWT ($\downarrow$) & Wiki ($\downarrow$)  \\ \hline
Softmax (GPT-2) & 125.0M & 82.9  & 18.96 & 24.28 & 355.9M & 207.8 & 15.81 & 20.12          \\
Softmax + Mi & 130.9M & 85.6  &18.74 & 24.08 & 366.4M & 213.8 & 15.71 & 20.07 \\
Mixture of Softmax (MoS)~\citep{yang2018breaking} & 126.2M & 130.2 & 18.97 & 24.10 & 358.0M & 262.9 & 15.71 & 19.95\\
MoS + Mi~\citep{chang2022softmax} & 133.3M & 133.2 & 18.68 & 23.82 & 370.6M & 268.2 & 15.61 & 19.86 \\
Pointer Generator (PG)~\citep{cnn_dataset_split} & 126.2M & 106.0 & 18.67 & 23.70 & 358.0M & 237.8 & 15.72 & 19.95\\
Pointer Sentinel (PS)~\citep{MerityX0S17} & 126.2M & 94.1  & 18.70 & 23.79 & 358.0M & 218.3 & 15.72 & 19.95\\
Softmax + R:20 + Mi &132.1M & 90.4  & 18.67 & 24.03 & 368.5M & 203.6 & 15.64 & 19.94 \\
Softmax + R:20,100 + Mi & 133.3M & 101.1 & 18.69 & 23.93 & 370.6M & 228.5 & 15.61 & 19.89 \\
Softmax + C + Mi & 132.1M & 94.8  & 18.48 & 23.56 & 368.5M & 222.7 & 15.60 & 19.83 \\
Softmax + P + Mi & 133.3M & 99.1  & 18.58 & 23.66 & 370.6M & 214.7 & 15.63 & 19.90 \\
PG + Mi & 133.3M & 111.2 & 18.43 & 23.43 & 370.6M & 242.5 & 15.60 & 19.89 \\
PS + Mi & 133.3M & 98.0  & 18.48 & 23.53 & 370.6M & 224.6 & 15.60 & 19.87 \\
Softmax + CR:20,100 + Mi & 134.5M & 113.3 & 18.46 & 23.48 & 372.7M & 234.5 & 15.54 & 19.75 \\
Softmax + CPR:20,100 + Mi & 136.8M & 119.9 & 18.43 & 23.42 & 376.9M & 249.9 & 15.53 & 19.71\\
MoS + CPR:20,100 + Mi & 139.2M & 165.1 & \textbf{18.39} & \textbf{23.29} & 381.1M & 300.6 & \textbf{15.44} & \textbf{19.57}\\
\end{tabular}
}
\caption{Comparison of different methods on top of GPT-2. Wiki and OWT refer to the testing perplexity of Wikipedia 2021 and OpenWebText, respectively. Lower perplexity is better. Time is the inference time of a batch; Mi is the multiple input hidden state enhancement; C is the context partition; R:20,100 is the reranker partition with $k_1=20$ and $k_2=100$; P is the pointer network (i.e., local decoder embedding). Please see \autoref{eq:logit_CPR} for the details of CPR.  The best scores are highlighted.
}
\label{tb:perplexity}
\end{table*}

\subsection{GPT-2}

We follow the setup in \citet{chang2022softmax} to continue training GPT-2 on Wikipedia 2021 and OpenWebText~\citep{radford2019language}. 

\subsubsection{Perplexity Comparison}

In \autoref{tb:perplexity}, we first compare their predictions on the next word distribution using the testing data perplexity, which is a standard metric in the LM architecture studies. In the table, Mi refers to multiple input state enhancement, which is proposed to break the softmax bottleneck more effectively (please see details in \Cref{sec:Mi} and \citet{chang2022softmax}). 

As we can see, \textbf{Softmax + CPR:20,100 + Mi}, which combines all the efficient approaches (i.e., context partition, reranker partition, and local decoder embedding), results in better performance and faster inference speed than the mixture of softmax (\textbf{MoS}) \citep{yang2018breaking,chang2022softmax}. The inference speed is measured by our pure PyTorch implementation, which we believe could be further accelerated by implementing some new PyTorch operations using CUDA code. 

If only using one method, the context partition (\textbf{Softmax + C + Mi}) is better than the reranker partitions (\textbf{Softmax + R:20,100 + Mi}) while performing similarly compared to local decoder word embedding (\textbf{Softmax + P + Mi}), Pointer Generator (\textbf{PG + Mi})~\citep{cnn_dataset_split}, and Pointer Sentinel (\textbf{PS + Mi})~\citep{MerityX0S17}.\footnote{Notice that the pointer networks from the previous work were originally designed for RNN. To add them on top of the transformer based LMs and make it more comparable to our methods, we simplify their architectures a little. Please see \Cref{sec:pointer_imp} for more details.} 
Their similar performances indicate that the improvement of pointer networks come from breaking the softmax bottleneck.
The significantly better performance of \textbf{PS + Mi} compared to \textbf{PS} further supports the finding.

To know how well our method breaks the softmax bottleneck, we implement a word-by-word reranker model on GPT-2, which appends the most likely 100 words to the context when predicting each next word (see \Cref{sec:wbw_rerank} for more details). 
In \autoref{tb:reranker}, we show that our efficient softmax alternative \textbf{Softmax + CPR:20,100 + Mi} achieves significantly lower perplexity. Furthermore, the word-by-word reranker is at least 10 times slower during training. Combining word-by-word reranker with our method only improves the perplexity very slightly, which suggests the challenges of further improving LM by breaking softmax bottleneck.





\begin{table}[t!]
\centering
\scalebox{0.68}{
\begin{tabular}{cc|cc}
\hline
Softmax + Mi &  29.33 & Softmax + wbwR:100 + Mi & 28.89 \\ \hline
Softmax + & \multirow{2}{*}{28.46} & Softmax +  &  \multirow{2}{*}{28.40} \\
CPR:20,100 + Mi &   & CPR:20,100 + wbwR:100 + Mi &  \\ \hline
\end{tabular}
}
\caption{Comparison between our method and word-by-word reranker for the most likely 100 words (wbwR:100). The numbers are the validation perplexities on Wikipedia 2021 after training for 0.15 epochs.}
\label{tb:reranker}
\end{table}

\begin{table}[t!]
\centering
\scalebox{0.7}{
\begin{tabular}{c|cc|cc|}
& \multicolumn{2}{c|}{All} & \multicolumn{2}{c|}{Proper Noun} \\
Model Name & Ref & Context & Ref & Context \\ \hline 
Softmax + Mi & 22.90          & 24.04          & 7.49          & 14.84 \\
MoS + Mi & 22.88          & 23.98          & 7.70          & 15.49 \\
PS + Mi & 22.85          & 25.01          & \textbf{8.16}          & 18.21 \\
Softmax + CPR:20,100 + Mi &  \textbf{23.05} & 25.36 & \textbf{8.16} & 17.92 \\ \hline


\end{tabular}
}
\caption{ROUGE-1 F1 (\%) of different methods on GPT-2. We compare the scores between the generated text and the reference (i.e., continuation), and between the generation and context. More methods and metrics are reported in \autoref{tb:gpt2_gen_other}. }
\label{tb:dynamic_partition_rouge_1}
\end{table}



\subsubsection{Generated Text Comparison}

Next, we would like to understand how the distribution improvement affects the text generation. We sample some contexts in the test set of Wikipedia 2021 and compare the generated text quality of the different models given the contexts. The quality is measured by the ROUGE-1 F1 scores between the generated text and the actual continuation. To know how much the different models copy from the context, we also report the ROUGE-1 scores between the generation and the contexts.

The results in \autoref{tb:dynamic_partition_rouge_1} show that different methods have very similar overall ROUGE-1 scores. Nevertheless, compared to \textbf{Softmax + Mi}, \textbf{Softmax + CPR:20,100 + Mi} is 21\% more likely to copy the proper nouns (i.e., entity names) from the context and 9\% more likely to generate the proper nouns in the actual continuation. This suggests that our method could alleviate the common incoherence problem of entities in generated text~\citep{shuster2021i,anonymous2022towards,zhang2022improving,guan2022generating,goyal2022snac}. In \autoref{tb:gpt2_gen_other}, we compare methods using more metrics to further support the conclusion.




%% file: content/analysis.tex
\begin{table*}[t!]
\centering
\scalebox{0.65}{
\begin{tabular}{|c|>{\centering\arraybackslash}p{5.5cm}>{\centering\arraybackslash}p{5.5cm}>{\centering\arraybackslash}p{6cm}|}
\hline
\multirow{3}{*}{Input Context} & There are plates, keys, scissors, toys, and balloons in front of me, and I pick up the & Choosing between John, Alex, Mary, Kathryn, and Jack, I decided to first talk to & I like tennis, baseball, golf, basketball, and \\ \hline \hline
\multirow{2}{*}{Softmax + Mi} & \textbf{keys}  0.108, \textbf{pieces}  0.045, key  0.036, phone  0.020, \textbf{balloons}  0.019 & \textbf{John}  0.108, the  0.102, them 0.095, him  0.045, my 0.032  & tennis  0.089, baseball  0.075, \textbf{football}  0.041, basketball  0.036, \textbf{I}  0.032 \\ \hline
\multirow{2}{*}{Mixture of Softmax (MoS) + Mi} & \textbf{keys}  0.085, phone  0.035, key  0.031, \textbf{pieces}  0.029, \textbf{balloons}  0.016 & \textbf{John}  0.099, the  0.097, them  0.083, \textbf{Alex}  0.055, \textbf{Mary}  0.040  & baseball  0.076, basketball  0.062, tennis  0.059, golf  0.037, \textbf{bad}  0.035 \\ \hline
\multirow{2}{*}{Pointer Sentinel (PS) + Mi}  & \textbf{keys}  0.091, \textbf{plates}  0.079, \textbf{scissors}  0.050, \textbf{balloons}  0.034, \textbf{toys}  0.033 & \textbf{John}  0.130, the  0.105,  \textbf{Alex}  0.076, them  0.076, \textbf{Mary}  0.037 & tennis  0.095, golf  0.050, baseball  0.043, \textbf{I}  0.038, \textbf{other} 0.038\\ \hline
\multirow{2}{*}{Softmax + CPR:20,100 + Mi}  & \textbf{keys}  0.077, \textbf{balloons}  0.052, \textbf{plates}  0.036, \textbf{toys}  0.030, \textbf{pieces}  0.030 & the  0.106, \textbf{John}  0.099,  my  0.060, \textbf{Alex}  0.057, them  0.044 & \textbf{football}  0.075, \textbf{volleyball}  0.058, \textbf{soccer}  0.056, \textbf{I}  0.047, \textbf{bad}  0.038\\ \hline

\end{tabular}
}
\caption{Prediction visualization of three input contexts. We show the top five words with the highest prediction probabilities of each model. The reasonable next word predictions are boldfaced.}
\label{tb:context_example}

\end{table*}

\begin{table*}[t!]
\centering
\scalebox{0.53}{
\begin{tabular}{|c|cccc|cccc|cccc|cccc|}
\hline
&  \multicolumn{4}{c|}{CNN/DM} & \multicolumn{4}{c|}{XSUM}&  \multicolumn{4}{c|}{BookSum Paragraph} & \multicolumn{4}{c|}{SAMSUM} \\
Model Name & R1 &  CIDEr  & factCC & MAUVE & R1 & CIDEr & factCC & MAUVE  & R1 & CIDEr & factCC & MAUVE  & R1 & CIDEr & factCC & MAUVE \\ \hline
\multicolumn{17}{|c|}{T5-Small} \\ \hline
Softmax (S) & 38.255          & 0.442          & 0.462          & 0.861          & 28.713          & 0.446          & 0.254          & 0.939          & 16.313          & 0.083          & 0.424          & 0.328          & 39.472          & 0.817          & 0.577          & 0.898          \\
CopyNet~\citep{gu2016incorporating}  & 37.990          & 0.438          & 0.482          & 0.865          & 28.573          & 0.442          & 0.274          & 0.940          & 16.666          & 0.092          & 0.439          & 0.402          & 39.525          & 0.853          & 0.579          & 0.924          \\
PG~\citep{cnn_dataset_split} & 37.913          & 0.442          & 0.467          & 0.874          & 28.777          & 0.450          & 0.257          & 0.931          & 16.432          & 0.088          & 0.429          & 0.376          & 32.451          & 0.585          & 0.552          & 0.153          \\
PS~\citep{MerityX0S17} & 38.058          & 0.444          & 0.466          & 0.854          & 28.442          & 0.435          & 0.267          & 0.932          & 16.408          & 0.090          & 0.436          & 0.395          & 38.731          & 0.817          & 0.578          & 0.865          \\
S + R:20 & 37.881          & 0.433          & 0.474          & 0.872          & 28.557          & 0.440          & 0.256          & 0.931          & 16.336          & 0.086          & 0.431          & 0.370          & 39.073          & 0.752          & 0.579          & 0.847          \\
S + E &  38.137          & 0.441          & 0.477          & 0.866          & 28.723          & 0.444          & 0.272          & 0.942          & 16.542          & 0.090          & 0.435          & 0.390          & 39.056          & 0.784          & 0.579          & 0.904          \\
S + CE & 38.461          & 0.460          & 0.475          & 0.874          & 29.155          & 0.464          & 0.270          & \textbf{0.948} & 16.628          & 0.093          & 0.436          & 0.403          & 40.055          & 0.835          & \textbf{0.583}          & 0.943          \\
S + CER:20 & 38.346          & 0.450          & \textbf{0.482} & \textbf{0.890} & 29.067          & 0.459          & \textbf{0.276} & 0.942          & 16.638          & 0.093          & 0.436          & 0.400          & \textbf{40.505} & 0.846          & 0.580          & 0.915          \\
S + CEPR:20 & \textbf{38.807} & \textbf{0.456} & 0.481          & 0.877          & \textbf{29.395} & \textbf{0.474} & 0.273          & 0.942          & \textbf{16.894} & \textbf{0.098} & \textbf{0.440} & 0.418          & 40.127          & \textbf{0.891} & 0.582          & \textbf{0.946} \\
S + CEPR:20 + Mi &  38.675          & 0.451          & 0.475          & 0.878          & 29.348          & 0.470          & 0.275          & 0.946          & 16.738          & 0.096          & 0.438          & \textbf{0.426} & 40.328          & 0.874          & 0.582 & 0.932         
 \\ \hline
\multicolumn{17}{|c|}{T5-Base} \\ \hline
Softmax (S) & 40.198          & 0.504          & 0.478          & 0.907          & 33.571          & 0.667          & 0.249          & 0.979          & 16.761          & 0.096          & 0.424          & 0.467          & 44.348          & 1.046          & 0.574          & \textbf{0.986} \\
CopyNet~\citep{gu2016incorporating}  & 39.940          & 0.507          & 0.484          & 0.903          & 33.557          & 0.666          & 0.253          & 0.979          & 16.918          & 0.101          & 0.430          & 0.531          & 44.141          & 1.052          & 0.570          & 0.973    \\
PG~\citep{cnn_dataset_split} & 39.982          & 0.489          & 0.485          & 0.911          & 33.605          & 0.663          & 0.255          & 0.982          & 16.611          & 0.095          & 0.423          & 0.463          & 37.597          & 0.784          & 0.548          & 0.140     \\
PS~\citep{MerityX0S17} &  40.018          & 0.495          & 0.483          & 0.914          & 33.638          & 0.672          & 0.249          & \textbf{0.983} & 16.905          & 0.100          & 0.428          & 0.504          & 43.098          & 1.008          & 0.575          & 0.946    \\
S + CEPR:20 & 40.354          & \textbf{0.511} & \textbf{0.487} & \textbf{0.919} & 33.700          & 0.675          & 0.260          & 0.980          & \textbf{16.997} & 0.100          & \textbf{0.432} & \textbf{0.549} & \textbf{44.860} & \textbf{1.064} & 0.573          & 0.963  \\
S + CEPR:20 + Mi &  \textbf{40.510} & 0.506          & 0.481          & 0.918          & \textbf{33.853} & \textbf{0.683} & \textbf{0.263} & \textbf{0.983} & 16.975          & \textbf{0.101} & 0.431          & 0.546          & 44.488          & 1.055          & \textbf{0.576} & 0.980   \\ \hline
\multicolumn{17}{|c|}{BART Base} \\ \hline
Softmax (S) & 39.390          & 0.428          & 0.479          & 0.900          & 35.675          & 0.814          & 0.241          & 0.985          & 16.393          & 0.094          & 0.414          & 0.404          & 45.132          & 1.129          & 0.567                                  & 0.966           \\
CopyNet~\citep{gu2016incorporating} & 39.385          & 0.438          & 0.484          & 0.906          & 35.515          & 0.814          & \textbf{0.251} & \textbf{0.988}          & 16.642          & \textbf{0.100} & \textbf{0.422} & \textbf{0.495} & 44.316          & 1.103          & 0.577          & 0.970 \\
PG~\citep{cnn_dataset_split} & 39.264          & 0.444          & 0.489          & \textbf{0.909} & 35.653          & 0.810          & 0.242          & 0.987          & 16.402          & 0.094          & 0.414          & 0.402          & \textbf{45.278} & 1.153          & \textbf{0.578} & 0.977  \\
PS~\citep{MerityX0S17} & 39.471          & \textbf{0.459} & \textbf{0.490} & 0.906          & 35.411          & 0.809          & 0.247          & 0.986          & \textbf{16.718} & 0.099          & \textbf{0.422} & 0.492          & 44.575          & 1.084          & 0.573          & 0.974       \\
S + R:20 & 39.181          & 0.434          & 0.475          & 0.905          & 35.586          & 0.808          & 0.247          & \textbf{0.988}          & 16.419          & 0.096          & 0.418          & 0.439          & 45.024                                  & \textbf{1.154} & 0.572                                  & 0.970  \\
S + E & 39.267          & 0.439          & 0.483          & 0.907          & 35.698          & 0.819          & 0.241          & \textbf{0.988} & 16.442          & 0.097          & 0.415          & 0.429          & 44.825                                  & 1.106                                  & 0.572                                  & 0.981         \\
S + CE &  39.416          & 0.442          & 0.481          & 0.908          & 35.727          & 0.812          & 0.241          & \textbf{0.988} & 16.555          & 0.096          & 0.417          & 0.435          & 44.295          & 1.116          & 0.572          & 0.985  \\
S + CER:20 & 39.421          & 0.439          & 0.482          & 0.900          & 35.576          & 0.812          & 0.236          & 0.987          & 16.553          & 0.096          & 0.418          & 0.454          & 45.054          & 1.150          & 0.576          & \textbf{0.988} \\
S + CEPR:20 & \textbf{39.723} & 0.441          & 0.483          & 0.908          & 35.732          & 0.822          & 0.242          & 0.986          & 16.664          & 0.098          & 0.420          & 0.467          & 44.732          & 1.115          & 0.575          & 0.974      \\
S + CEPR:20 + Mi &  39.626          & 0.442          & 0.482          & 0.907          & \textbf{35.846} & \textbf{0.828} & 0.245          & 0.986          & 16.597          & 0.097          & 0.419          & 0.466          & 44.728          & 1.132          & 0.574          & \textbf{0.988} \\ \hline
\multicolumn{17}{|c|}{BART Large} \\ \hline
Softmax (S) & 40.749          & 0.424          & 0.495          & 0.899          & 38.828          & 0.921          & \textbf{0.263} & 0.988          & 17.271          & 0.103          & 0.420          & 0.461 & 47.384          & 1.187          & \textbf{0.574} & 0.975 \\
CopyNet~\citep{gu2016incorporating}  & 40.622          & 0.407          & 0.487          & 0.890          & 38.576          & 0.920          & 0.258          & 0.989          & 17.342          & \textbf{0.106} & 0.425          & 0.512 & 47.911 & 1.232          & 0.573          & 0.980        \\
PG~\citep{cnn_dataset_split} & 40.766          & 0.407          & 0.489          & 0.902          & 38.869          & 0.944          & 0.256          & 0.990          & 17.289          & 0.103          & 0.424          &  0.470& 47.737          & 1.199          & 0.573          & 0.964 \\
PS~\citep{MerityX0S17} & 40.643          & 0.424          & \textbf{0.502}          & 0.907          & 38.886          & 0.952          & 0.255          & 0.988          & \textbf{17.382} & 0.105          & \textbf{0.426} & \textbf{0.527} & \textbf{48.253}          & 1.246          & \textbf{0.574}          & \textbf{0.986} \\
S + CEPR:20 & \textbf{40.876} & 0.458          & 0.500 & 0.925          & \textbf{38.991} & 0.955          & 0.248          & 0.990          & 17.337          & \textbf{0.106}          & 0.423          & 0.467 & 47.253          & \textbf{1.298} & 0.572          & 0.976    \\
S + CEPR:20 + Mi &  40.441          & \textbf{0.463} & 0.500          & \textbf{0.927} & 38.705          & \textbf{0.965} & 0.242          & \textbf{0.991} & 16.995          & 0.105          & 0.421          &0.482  & 47.488          & 1.271          & 0.571          & \textbf{0.986}  \\ \hline
\end{tabular}
}
\caption{The performance on test sets of four summarization datasets. R1 is ROUGE-1 F1 (\%). E refers to the encoder partition; C is the context partition; R:20 is the reranker partition with $k_1=20$; The P in CEPR means using the pointer networks for both encoder (LE) and decoder (LD); Mi is the multiple input hidden state enhancement; PS means Pointer Sentinel and PG means Pointer Generator. CEPR is described in \autoref{eq:logit_CEPR}. The model size, inference time, and more metrics are reported in \autoref{tb:summarization_other_news} and \autoref{tb:summarization_other_types}. 
}
\label{tb:summarization}
\end{table*}

\subsubsection{Qualitative Analysis}


In \autoref{tb:context_example}, we visualize some distributions to explain our improvements.
The softmax layer of GPT-2 is unable to properly learn to copy or exclude the word from the input context. For example, \textbf{Softmax + Mi} and \textbf{MoS + Mi} might output ``\textit{There are plates, keys, scissors, toys, and balloons in front of me, and I pick up the phone}'', which causes a hallucination problem, while \textbf{Softmax + CPR:20,100 + Mi} and \textbf{Pointer Sentinel (PS) + Mi} can output the mentioned options with similar probability by copying the words in the context. In addition, \textbf{GPT-2}, \textbf{MoS}, and \textbf{PS + Mi} are very likely to output ``\textit{I like tennis, baseball, golf, basketball, and tennis}''. This repetition problem happens because the next word should be some words similar to the listed sports names except for the sports that have been mentioned and the softmax layer has difficulties in outputting a donut-shape next word distribution in embedding space. 
In contrast, \textbf{Softmax + CPR:20,100 + Mi} can learn to exclude the listed sports by putting very negative logits on the context words, which yield the desired donut-shape distribution. 



%% file: content/experiments_ED.tex
\subsection{T5 and BART in Summarization}

We test our methods on two popular encoder-decoder LMs, T5~\citep{2020t5} and BART~\citep{lewis2020bart}. We fine-tune the pretrained LMs with different softmax alternatives on two news summarization datasets: CNN/DM~\citep{cnn_dataset_split} and XSUM~\citep{narayan2018don}, one narrative summarization dataset: BookSum at paragraph level~\citep{kryscinski2021booksum}, and one dialogue summarization dataset: SAMSUM~\citep{gliwa-etal-2019-samsum}.

In the main paper, we evaluate the quality of summaries using four metrics. ROUGE-1 F1~\citep{lin2004rouge} measures the unigram overlapping between the generated summary and the ground truth summary; CIDEr~\citep{vedantam2015cider} adds a tf-idf weighting on the n-gram overlapping score to emphasize correct prediction of rare phrases; factCC~\citep{kryscinskiFactCC2019} evaluates the factuality of the summary; MAUVE~\citep{pillutla2021mauve} compares the word distribution of summary and ground truth in a quantized embedding space. To further support our conclusions, we also compare the quality measured by several other metrics and their model sizes in \autoref{tb:summarization_other_news} and 
\autoref{tb:summarization_other_types}.

The results are reported in
\autoref{tb:summarization}. Similar to the GPT-2 experiments, the results are generally better as we combine more partitions and local embedding approaches. This demonstrates that we can directly fine-tune the LMs with our softmax alternatives without expensive pretraining.

Unlike the GPT-2 experiments, multiple input hidden state enhancement (Mi) is not very effective, so we mainly compare the methods without Mi (i.e., $\vq_{c_t}=\vh_{c_t}^M$, unlike \autoref{eq:multi-hidden}). We hypothesize one possible reason is that we haven't pretrained the T5 and BART with our softmax alternatives.  

Our improvements are larger in smaller models. This is probably because in a smaller word embedding space, there are more likely to be interfering words between the desired next word possibilities.
Compared to our methods, the pointer networks perform well in BART-base but usually perform worse in other LMs. We need further investigations in the future to explore the reasons.

Compared to ROUGE-1 score, the improvement percentage of CIDEr is overall higher. One major problem of the summarization LMs is that the generated summary contains too many commonly used phrases~\citep{king2022don} and our considerably higher CIDEr scores indicate the alleviation of the problem. Our improvement on the factCC is also significant~\citep{cao2021cliff}. 
Finally, our MAUVE improvement percentage on BookSum Paragraph dataset could reach around 30\% in T5-Small. We hypothesize this is because we often mention the global entity names in the news (e.g., Obama) while the meaning of names in stories (e.g., John) is often defined by the context.





%% file: content/related_work.tex
\section{Related Work}

Repetition and hallucination are two common problems in language generation tasks. One common solution for repetition is to avoid outputting the words in the context, which is often called unlikelihood training~\citep{welleck2019neural, jiang2022simple, su2022contrastive}. However, when LM should mention some names in the context, this might exacerbate the hallucination problem. In contrast, our method can learn to copy and exclude the words in context as in \autoref{tb:context_example}.

To alleviate the hallucination problem or satisfy some constraints, many recent generation models rerank the generated text~\citep{deng2020residual,gabriel2021discourse,cobbe2021training,ravaut2022summareranker,krishna2022rankgen,glass2022re2g,an2022cont,arora2022director,adolphs2022cringe,meng2022controllable,mireshghallah2022mix,kumar2022gradient,wan2022factpegasus,jiang2022pairreranker}. Although being effective, the rerankers usually slow down significantly the training and/or inference speed (as our word-by-word reranker baseline) and might occupy extra memory resources.

Our analyses demonstrate that parts of the hallucination and repetition problem come from the softmax bottleneck. The findings provide an explanation for the effectiveness of prior studies such as the above reranker approaches and pointer networks~\citep{li2021learn,ZhongLC22,ma2023prom}. Another example is encouraging the word embeddings to be isotropy~\citep{wang2020improving,su2022contrastive}. Their improvement might also come from reducing linear dependency of the candidate word embeddings. Nevertheless, their side effect of breaking the similarity structure in the word embedding space might hurt the generation quality in some cases. Concurrently to our work, \citet{wan2023histalign} also use the softmax bottleneck theory~\citep{chang2022softmax} to explain the improvement of a pointer network. Their empirical results also support our conclusion that softmax bottleneck is a major reason that causes the factuality problem of LMs.

Our work is motivated and inspired by \citet{chang2022softmax}. In their work, they also propose to use different hidden states for different vocabulary partitions, but their partitioning is global and needs to be combined with the mixture of softmax (MoS) approach, which adds a significant overhead compared to the standard softmax layer. Our dynamic partitioning methods not only perform better but greatly reduce the overhead by removing the reliance on MoS. 

%% file: content/ethics.tex
\begin{figure*}[t!]
\centering
\begin{subfigure}{0.49\textwidth}
  \centering
  \includegraphics[width=1\linewidth]{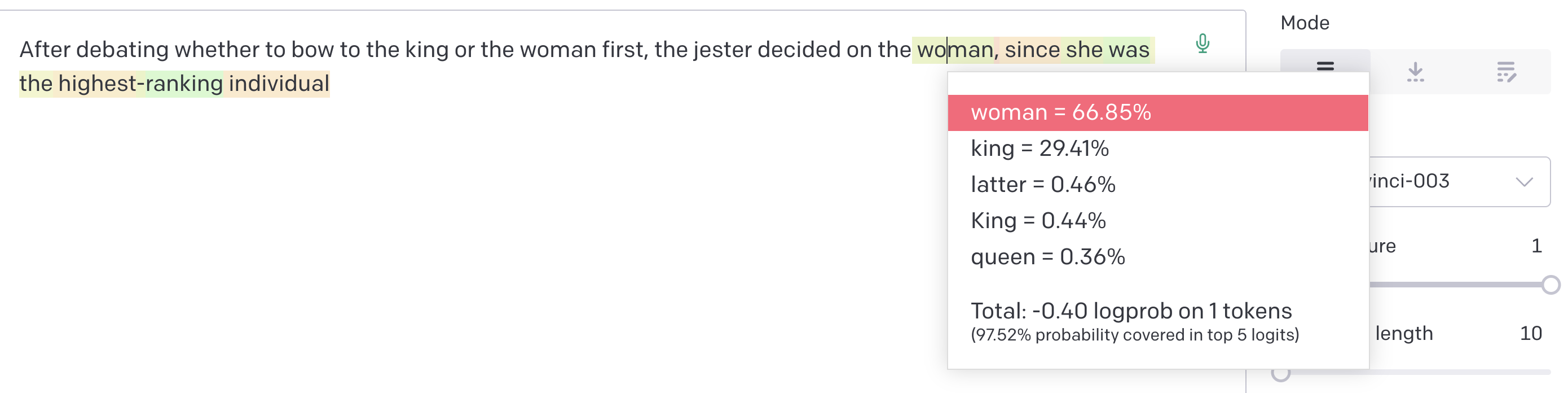}
  \caption{The example where the next word should be either \textit{woman} or \textit{king} (or their synonym such as former and latter). }
  \label{fig:gpt-3_king}
\end{subfigure} \;\;%
\begin{subfigure}{0.49\textwidth}
  \centering
  \includegraphics[width=1\linewidth]{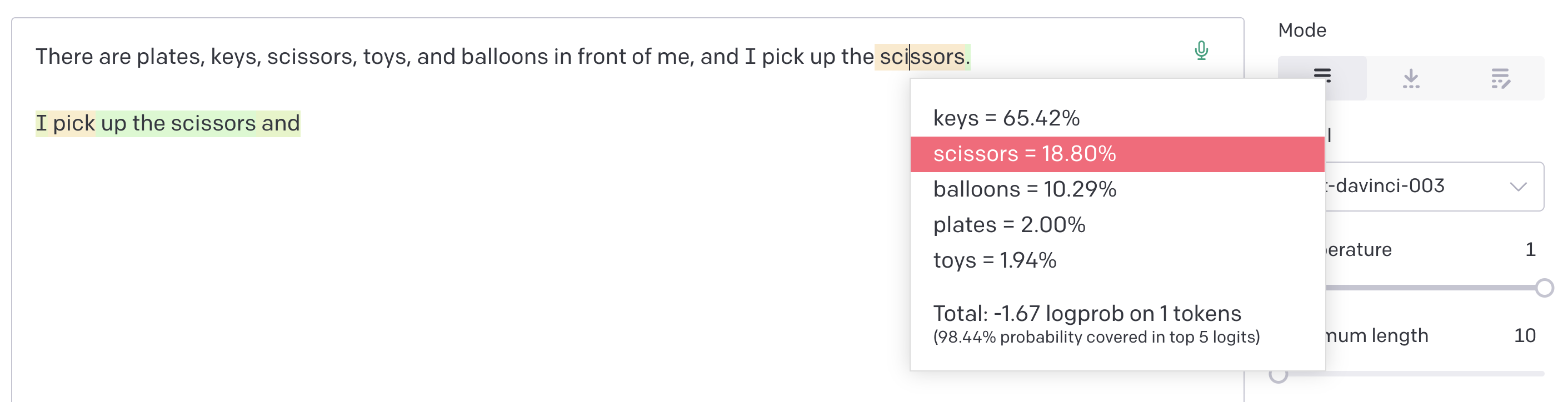}
  \caption{The example where the next word \textit{plates}, \textit{keys}, \textit{scissors}, \textit{toys}, and \textit{balloons} should receive similar probabilities.}
  \label{fig:gpt-3_key}
\end{subfigure}
\begin{subfigure}{0.49\textwidth}
  \centering
  \includegraphics[width=1\linewidth]{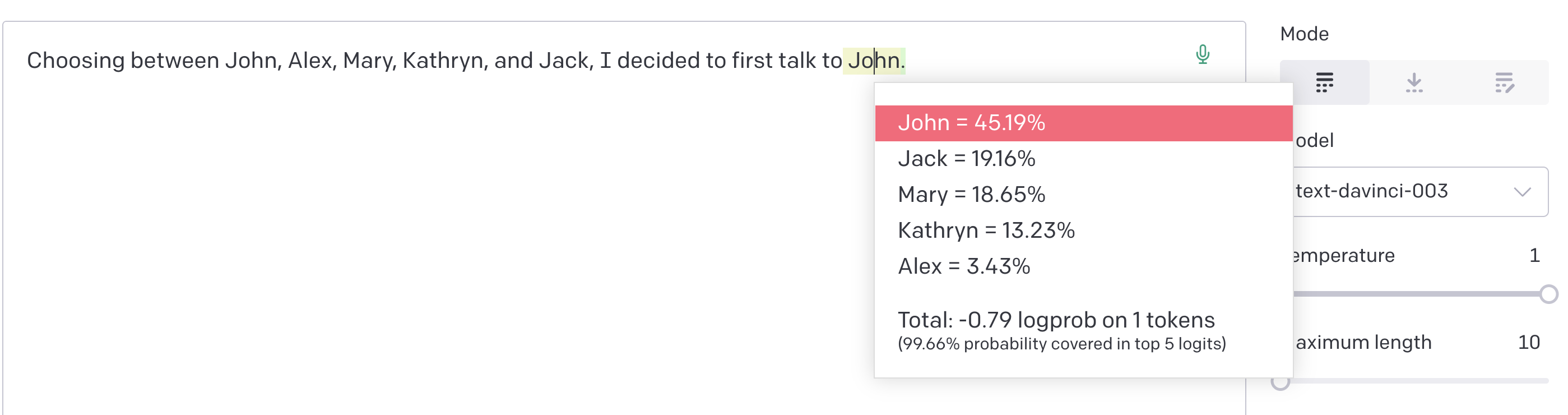}
  \caption{The example where the next word \textit{John}, \textit{Alex}, \textit{Mary}, \textit{Kathryn}, and \textit{Jack} should receive similar probabilities.}
  \label{fig:gpt-3_John}
\end{subfigure} \;\;%
\begin{subfigure}{0.49\textwidth}
  \centering
  \includegraphics[width=1\linewidth]{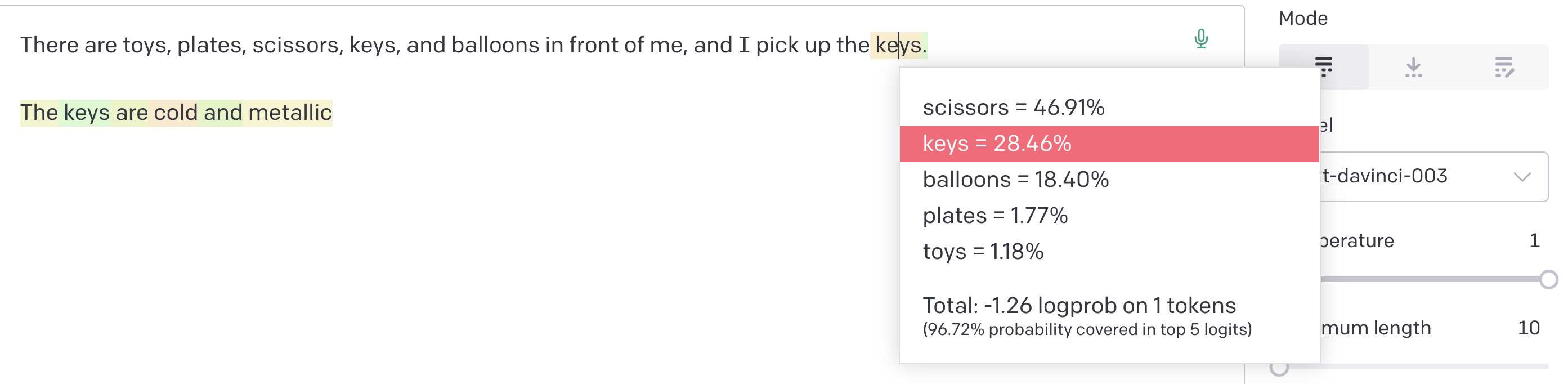}
  \caption{Same as above except that the order of the objects in the context is different.}
  \label{fig:gpt-3_key2}
\end{subfigure}
\caption{The next word probabilities outputted by GPT-3.5 (text-davinci-003).}
\label{fig:gpt-3}
\end{figure*}

\section{Limitations}

In our experiments, we find that the improvement of our methods tend to be larger in relatively smaller language models. Due to our limited access of computational resources, we are not able to try our methods on larger LMs. To know if a larger LM still suffers from the softmax bottleneck problem, we input the examples we used in \autoref{tb:context_example} to GPT-3.5 and report their results in \autoref{fig:gpt-3}. 

We find that although GPT-3.5 greatly reduces the chance of hallucination compared to GPT-2, the next word distribution is still not ideal. For example, in \autoref{fig:gpt-3_king}, although the incorrect answer \textit{queen} receives only a small probability, GPT-3.5 puts around 67\% probability on \textit{woman}. Similarly, even though GPT-3.5 is unlikely to hallucinate the sentence: \textit{There are plates, keys, scissors, toys, and balloons in front of me, and I pick up the phone} as GPT-2, \autoref{fig:gpt-3_key} and \autoref{fig:gpt-3_key2} show that the output distribution is still heavily biased toward one of the options and the most likely next word could change if the order of the options in the context changes. These results suggest that increasing model size indeed alleviates the softmax bottleneck problem but the problem is not completely solved even if a huge hidden state size (12k) and model size (175B) are used~\citep{brown2020language}. We expect that adding our methods to the large LMs could rectify the biased distributions as shown in our experiments on smaller LMs (\autoref{tb:context_example}). Therefore, although improving smaller LMs has already had wide applications in practice, trying our methods on a larger LM is a promising next step, which we haven't been able to do.

The current implementation of our methods also has some room for improvements. Our codes currently contain some unnecessary computation to circumvent the restrictions of PyTorch library, so we should be able to further accelerate it by writing CUDA code. Furthermore, our codes haven't supported the pretraining of BART or T5. We expect that completing the future work could make our method faster and better. 

Since the focus of this paper is improving the architecture of general transformer decoder, our evaluation of each application is not as comprehensive as the studies for a particular application. For example, although we test our methods using many metrics and the metrics show a consistent trend, there are many other factuality metrics we haven't tried~\citep{li2022faithfulness}. We also haven't conducted human evaluation to further verify our conclusion because conducting human evaluation properly is challenging~\citep{karpinska2021perils} and time-consuming. In addition, if we include more words in a context partition, the performance might be better at the cost of extra computational overhead. We leave the analyses of the tradeoff as future work.

\section{Ethics Statement}
In our experiments, we find that our methods usually copy more words from the context or encoder input. The tendency might have some potential issues. For example, our improvements might be reduced on the languages with more morphology. Furthermore, in some summarization applications, increasing the factuality by increasing the extractiveness might not be ideal~\citep{ladhak2022faithful,goyal2022news}.

As described in \Cref{sec:bottleneck}, one major limitation of the popular softmax layer is its global word embeddings. The problem would become more serious when there are more tokens whose meanings are locally defined (e.g., names in the BookSum dataset). Our methods would be more useful in those circumstances and might alleviate some biases described in \citet{shwartz2020you} and \citet{ladhak2023pre}. Moreover, the meaning of tokens are also locally defined in many other applications such as variables in code or math problems, the new terminologies in a scientific paper, or the products in a sequential recommendation problem. We believe that our methods could become an efficient alternative of reranker~\citep{cobbe2021training,welleck2022naturalprover} and create impacts in those areas.

Finally, our results show that when there are some uncertainties in the next word (e.g., could be \textit{king} or \textit{woman}), existing LMs could have some difficulties of copying the words from the context and our methods alleviate the problem. Thus, our methods should also be able to improve the lexically controllable language generation models that put the desired keywords into the context such as \citet{goldfarb2019plan} and \citet{lu2021neurologic}.

%% file: appendix/more_exp.tex
\begin{table*}[t!]
\centering
\scalebox{0.73}{
\begin{tabular}{|c|c|cccc|c|cccc|}
\hline
& \multicolumn{5}{c|}{Diagonal (e.g., \emph{king} or \emph{woman})} & \multicolumn{5}{c|}{Edge (e.g., \emph{king} or \emph{queen})} \\ \cline{2-11}
Analogy Relation Types $\rightarrow$ &  & capital- & capital- & city-in- & \multirow{2}{*}{family} &  & capital- & capital- & city-in- & \multirow{2}{*}{family} \\
Models $\downarrow$ & valid & common & world & state & & valid & common & world & state & \\ \hline
Softmax + Mi &  2.27 & 3.36 & 1.94 & 2.32 & 3.09 & 2.13 & 2.61 & 1.90 & 2.21 & 2.49  \\
MoS + Mi~\citep{chang2022softmax}  & 1.86 & 2.62 & 1.66 & 1.86 & 3.59 & 1.87 & 2.24 & 1.66 & 1.90 & 3.10\\
Softmax + C + Mi & 1.78 & 2.19 & 1.62 & 1.87 & 2.17 & 1.79 & 2.13 & 1.63 & 1.88 & 2.07\\
Softmax + CPR:20,100 + Mi  & \textbf{1.69} & \textbf{2.03} & \textbf{1.54} & \textbf{1.81} & \textbf{2.09} & \textbf{1.69} & \textbf{2.01} & \textbf{1.55} & \textbf{1.81} & \textbf{1.97} \\ \hline
\end{tabular}
}
\caption{ Comparing the perplexity of different \emph{GPT-2 Small} models using the synthetic dataset from \citet{chang2022softmax}.} 
\label{tb:analogy}
\end{table*}

\section{More Results and Analysis}
\label{sec:more_results}
In this section, we will report more results and provide more detailed analyses accordingly to investigate the advantages of different methods. 


\begin{table*}[t!]
\centering
\scalebox{0.8}{
\begin{tabular}{c|cccccccc}

Model Name & R1 & R1C & R1P & R1PC & R2 & P Ratio & CIDEr & NIST \\ \hline
Softmax (GPT-2) & 22.668          & 23.548          & 7.323          & 14.340          & 3.219          & 0.885 & 0.182          & 1.792 \\ 
Softmax + Mi & 22.903          & 24.036          & 7.493          & 14.840          & 3.289          & 0.877 & 0.190          & 1.829  \\
Mixture of Softmax (MoS)~\citep{yang2018breaking} & 22.965          & 24.233          & 7.760          & 15.762          & 3.260          & 0.885 & 0.188          & 1.846 \\
MoS + Mi~\citep{chang2022softmax} & 22.876          & 23.979          & 7.703          & 15.493          & 3.270          & 0.889 & 0.188          & 1.829 \\
Pointer Generator (PG)~\citep{cnn_dataset_split} & 23.055          & 24.872          & 8.052          & 17.830          & 3.311          & 0.889 & 0.193          & 1.856 \\
Pointer Sentinel (PS)~\citep{MerityX0S17} & 23.007          & 24.444          & 7.677          & 16.146          & 3.302          & 0.873 & 0.189          & 1.840 \\
Softmax + R:20 + Mi & 22.941          & 23.970          & 7.467          & 14.733          & 3.303          & 0.896 & 0.188          & 1.833 \\
Softmax + R:20,100 + Mi & 22.909          & 23.938          & 7.537          & 15.066          & 3.280          & 0.870 & 0.190          & 1.829 \\
Softmax + C + Mi & \textbf{23.116} & 25.027          & 7.894          & 17.048          & \textbf{3.372} & 0.917 & 0.197          & \textbf{1.873} \\
Softmax + P + Mi & 23.015          & 25.080          & 7.895          & 17.184          & 3.346          & 0.877 & 0.196          & 1.847 \\
PG + Mi & 22.827          & 24.759          & 8.049          & 17.874          & 3.289          & 0.914 & 0.191          & 1.819 \\
PS + Mi & 22.846          & 25.008          & 8.159          & 18.208 & 3.307          & \textbf{0.921} & 0.194          & 1.823 \\
Softmax + CR:20,100 + Mi & 23.017          & 25.056          & 8.089          & 17.798          & 3.328          & 0.894 & \textbf{0.198}          & 1.858 \\
Softmax + CPR:20,100 + Mi & 23.053          & 25.361 & 8.160          & 17.921          & 3.363          & 0.882 & 0.197          & 1.863 \\
MoS + CPR:20,100 + Mi & 23.047          & 25.173          & \textbf{8.187} & 18.198          & 3.314          & 0.902 & \textbf{0.198} & 1.868 \\
\end{tabular}
}
\caption{Comparison of the continuation generated by GPT-2 Small in Wikipedia test data. \autoref{tb:dynamic_partition_rouge_1} is a short summary of this table. The meaning of the metrics is described in \Cref{sec:gpt-2_more_results}. Higher R1C and R1PC mean copying more words from the context. A higher P Ratio means generating more proper nouns. All ROUGE scores are percentages.}
\label{tb:gpt2_gen_other}
\end{table*}

\begin{figure}[t!]
\centering
\includegraphics[width=1\linewidth]{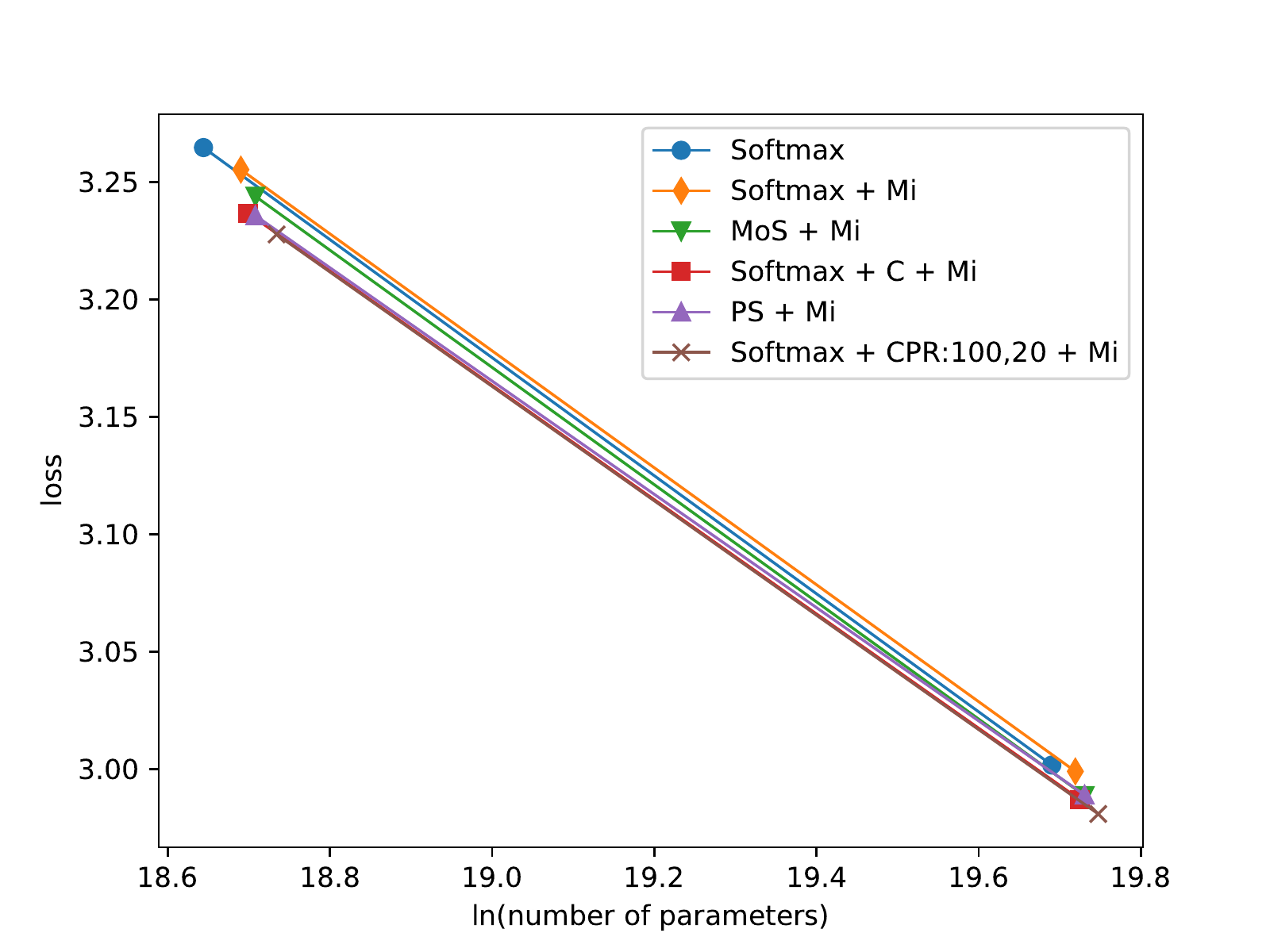}
\caption{The model size versus the model loss in Wikipedia test data after training for 0.4 epochs. The left side points are the results from GPT-2 Small and the right side points come from GPT-2 Medium. The lower curves are better.}
\label{fig:gpt_loss_size}
\end{figure}

\subsection{GPT-2 Experiments}
\label{sec:gpt-2_more_results}

\citet{kaplan2020scaling, henighan2020scaling} demonstrate that the loss decreases linearly as the log of the model size increases. Therefore, a new architecture needs to perform better than the old architecture with a similar model size to verify that the improvement does not come from memorizing more information through the extra parameters. From the 
loss versus log(model size) curve in \autoref{fig:gpt_loss_size}, we can see that our proposed methods are significantly better than MoS and slightly better than a pointer network baseline as the model becomes larger.

We use the following metrics to measure the text generated by GPT-2.
\begin{itemize}[leftmargin=.2in,topsep=0pt]
\setlength\itemsep{0.0em}
\item \textbf{ROUGE-1 (R1)}: The prediction F1 for unigram in the actual continuation.
\item \textbf{ROUGE-1 Context (R1C)}: The prediction F1 for unigram in the context.
\item \textbf{ROUGE-1 Proper (R1P)}: The same as ROUGE-1 except that only the proper nouns are considered. We measure this metric because the correctness of the entity name prediction is critical to the factuality of the generation.
\item \textbf{ROUGE-1 Proper Context (R1PC)}: The same as ROUGE-1 Context (R1C) except that only the proper nouns are considered.
\item \textbf{ROUGE-2 (R2)}: The prediction F1 for bigram in the actual continuation.
\item \textbf{Proper Noun Ratio (P Ratio)}: The average number of proper nouns in the generation divided by the average number of proper nouns in the actual continuation. The LMs usually generate fewer proper nouns compared to the actual continuation~\citep{see2019massively}, so the values are usually lower than 1. The P Ratio closer to 1 is better. 
\item \textbf{CIDEr~\citep{vedantam2015cider}}: A metric for measuring the quality and specificity of the generation.
\item \textbf{NIST~\citep{doddington2002automatic}}: Similar to CIDEr. CIDEr uses tf-idf to weigh the n-gram while NIST measures the information gain. 
\end{itemize}

\begin{figure}[t!]
\centering
\includegraphics[width=1\linewidth]{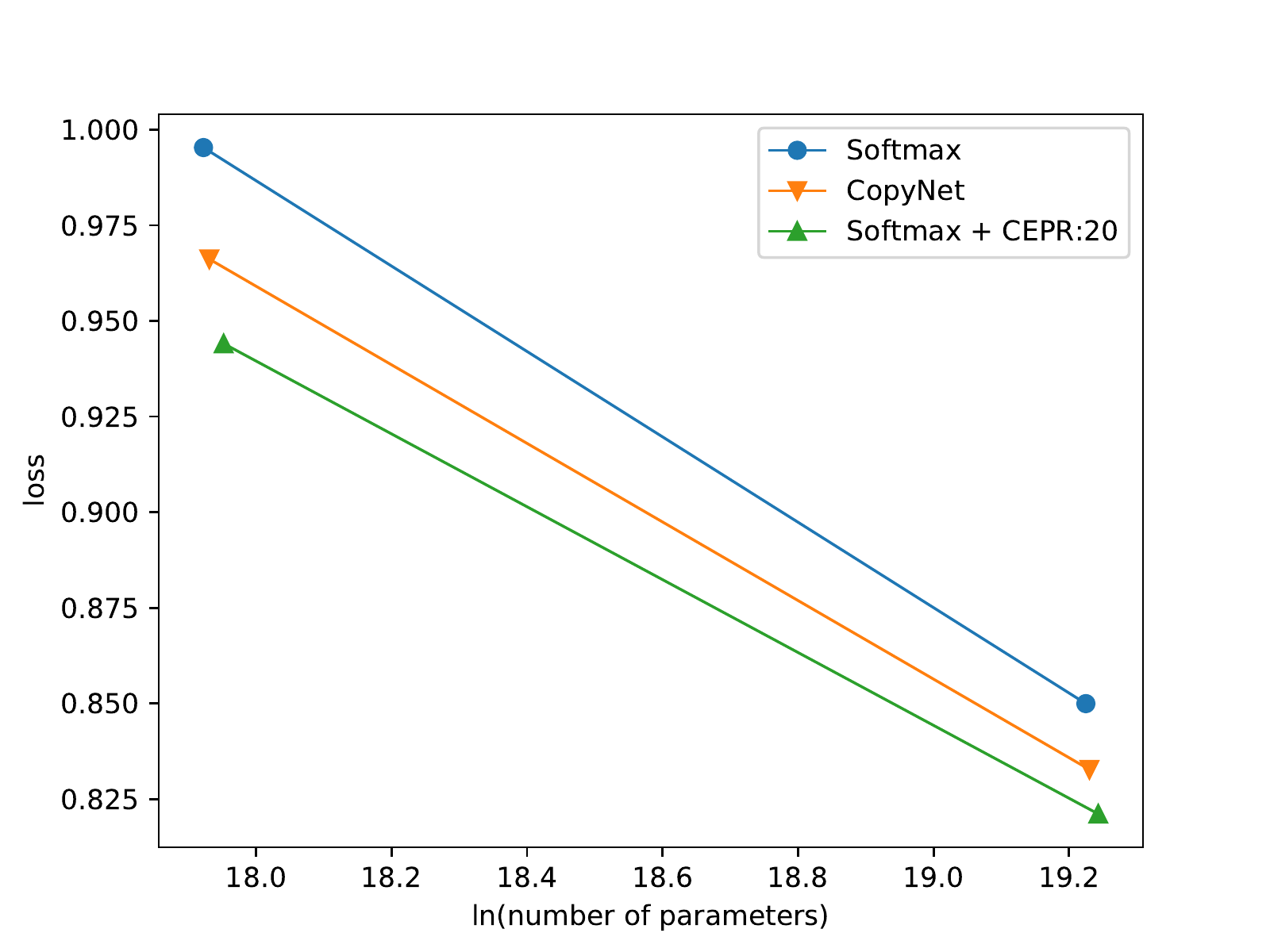}
\caption{The model size versus the model loss in CNN/DM test set. The left side points are the results from T5-Small and the right side points come from T5-Base. The lower curves are better.}
\label{fig:t5_loss_size}
\end{figure}

The results are reported in \autoref{tb:gpt2_gen_other}. In terms of R1, R2, CIDEr, and NIST, our proposed methods such as \textbf{Softmax + C + Mi} and \textbf{Softmax + CPR:20,100 + Mi} are significantly better than the pointer network baselines \textbf{PS + Mi} and \textbf{PG + Mi}. Comparing with \textbf{Softmax + CPR:20,100 + Mi}, \textbf{PS + Mi} has a significantly higher P Ratio and R1PC but similar R1P. This indicates that \textbf{PS + Mi} copies more proper nouns from the context while there is a similar number of proper nouns that are in actual continuation, so \textbf{Softmax + CPR:20,100 + Mi} actually has a higher accuracy on the proper noun prediction. 

In text corpus such as Wikipedia, we do not know the ground truth next word distribution and which context leads to multiple probable next words, so we cannot quantitatively analyze the improvement on the ambiguous contexts. To alleviate the concern, we test our methods on the synthetic dataset constructed by \citet{chang2022softmax}. The dataset is built using templates and Google analogy dataset~\citep{mikolov2013distributed}, so we know the ground truth next word distribution. The dataset consists of the ambiguous contexts such as \textit{I went to Paris and Germany before, and I love one of the places more, which is}, where the next word is either the diagonal words of the parallelogram such as \textit{Paris} and \textit{Germany} or the edge words such as \textit{Paris} and \textit{France}. For the details of the experimental setup, please refer to \citet{chang2022softmax}.

In \autoref{tb:analogy}, we can see that \textbf{Softmax + CPR:20,100 + Mi} achieves the lowest perplexity in all subsets and outperforms the \textbf{Softmax + Mi} baseline by a large margin, especially in the diagonal subset where the ground truth word embedding distribution has multiple modes. Notice that the performance of \textbf{MoS + Mi} is worse than what reported in \citet{chang2022softmax} probably because we shared the input and output word embeddings.






\begin{table*}[t!]
\centering
\scalebox{0.75}{
\begin{tabular}{c|c|ccccc|ccccc|}
&  & \multicolumn{5}{c|}{CNN/DM} & \multicolumn{5}{c|}{XSUM} \\
Models & Size & Loss ($\downarrow$) & R2 & R1P & P Ratio & NIST  & Loss ($\downarrow$) & R2 & R1P & P Ratio & NIST \\ \hline
\multicolumn{12}{|c|}{T5-Small} \\ \hline
Softmax (S) & 60.8M & 0.995          & 15.147          & 0.462          & 0.915          & 4.650          & 0.538          & 7.098          & 0.292          & 0.853          & 2.738          \\
CopyNet~\citep{gu2016incorporating} & 61.3M & 0.966          & 14.942          & 0.458          & \textbf{0.985} & 4.607          & 0.533          & 7.055          & 0.286          & 0.865          & 2.742          \\
PG~\citep{cnn_dataset_split} & 61.3M & 0.978          & 14.789          & 0.453          & 0.943 & 4.589          & 0.535          & 7.211          & 0.288          & 0.849          & 2.744          \\
PS~\citep{MerityX0S17} & 61.3M & 0.970 & 14.866 & 0.455 & 0.946          & 4.629          & 0.535          & 7.000          & 0.283          & 0.853          & 2.718          \\
S + R:20 & 61.0M  &0.985          & 14.831          & 0.456          & 0.928          & 4.603          & 0.534          & 7.085          & 0.287          & 0.858          & 2.730          \\
S + E & 61.0M & 0.956          & 14.935          & 0.457          & 0.950          & 4.629          & 0.530          & 7.152          & 0.292          & 0.864          & 2.759          \\
S + CE &  61.3M & 0.954          & 15.124          & 0.462          & 0.956          & 4.691          & 0.528          & 7.304          & 0.297          & 0.873          & 2.815          \\
S + CER:20 & 61.5M & 0.953          & 14.996          & 0.463          & 0.953          & 4.667          & 0.527          & 7.194          & 0.296          & 0.871          & 2.800          \\
S + CEPR:20 & 62.6M & 0.944          & \textbf{15.194} & \textbf{0.476} & 0.971          & \textbf{4.739} & 0.525          & \textbf{7.363} & \textbf{0.305} & \textbf{0.878} & \textbf{2.844} \\
S + CEPR:20 + Mi & 65.5M & \textbf{0.943} & 15.094          & 0.471          & 0.976          & 4.720          & \textbf{0.523} & 7.340          & \textbf{0.305} & 0.874          & 2.840   \\ \hline
\multicolumn{12}{|c|}{T5-Base} \\ \hline
Softmax (S) & 223.5M & 0.850          & 16.410          & 0.491          & 0.959          & 4.948          & 0.417          & 10.773          & 0.386          & 0.910          & 3.454 \\
CopyNet~\citep{gu2016incorporating}  & 224.7M & 0.833          & 16.253          & 0.486          & 0.979          & 4.915          & 0.416          & 10.804          & 0.387          & 0.915          & 3.467 \\
PG~\citep{cnn_dataset_split} &  224.7M & 0.840          & 16.134          & 0.485          & 0.955          & 4.923          & 0.417          & 10.815          & 0.389          & 0.915          & 3.466 \\
PS~\citep{MerityX0S17} & 224.7M & 0.836          & 16.275          & 0.490          & 0.978          & 4.908          & 0.417          & 10.838          & 0.386          & 0.915          & 3.473\\
S + CEPR:20 &  227.6M & \textbf{0.821} & 16.292          & 0.497          & \textbf{0.990} & 4.966          & \textbf{0.412} & 10.778          & 0.389          & \textbf{0.930} & 3.477  \\
S + CEPR:20 + Mi & 234.1M & \textbf{0.821} & \textbf{16.457} & \textbf{0.499} & 0.987          & \textbf{4.997} & \textbf{0.412} & \textbf{10.921} & \textbf{0.391} & 0.929          & \textbf{3.511} \\ \hline
\multicolumn{12}{|c|}{BART Base} \\ \hline
Softmax (S) & 140.0M & 0.874          & 15.613          & 0.471          & 1.028          & 4.641          & 0.391          & 12.944          & \textbf{0.428} & 0.928          & 3.833    \\
CopyNet~\citep{gu2016incorporating} &   141.2M & \textbf{0.837} & 15.675          & 0.470          & 1.013          & 4.685          & \textbf{0.387} & 12.740          & 0.424          & 0.934          & 3.818   \\
PG~\citep{cnn_dataset_split} & 141.2M & 0.845          & 15.485          & 0.465          & 1.018          & 4.669          & 0.389          & 12.849          & 0.425          & 0.928          & 3.827  \\
PS~\citep{MerityX0S17} & 141.2M & 0.838          & 15.689          & 0.468          & \textbf{0.996} & \textbf{4.750} & \textbf{0.387} & 12.690          & 0.423          & 0.926          & 3.796\\
S + R:20 & 140.6M & 0.863          & 15.486          & 0.468          & 1.028          & 4.655          & 0.389          & 12.804          & 0.426          & 0.941          & 3.824     \\
S + E & 140.6M & 0.852          & 15.412          & 0.466          & 1.018          & 4.652          & 0.389          & 12.893          & \textbf{0.428} & 0.933          & 3.844     \\
S + CE & 141.2M & 0.851          & 15.555          & 0.471          & 1.013          & 4.692          & 0.388          & 12.830          & 0.426          & 0.934          & 3.827     \\
S + CER:20 & 141.8M &  0.850          & 15.550          & 0.469          & 1.022          & 4.672          & \textbf{0.387} & 12.787          & 0.423          & 0.940          & 3.821 \\
S + CEPR:20 & 144.1M &  0.841          & \textbf{15.778} & 0.471          & 1.025          & 4.724          & \textbf{0.387} & 12.824          & 0.423          & \textbf{0.942} & 3.829     \\
S + CEPR:20 + Mi & 150.6M & 0.843          & 15.632          & \textbf{0.472} & 1.027          & 4.700          & \textbf{0.387} & \textbf{12.969} & 0.426          & 0.939          & \textbf{3.847} \\ \hline
\multicolumn{12}{|c|}{BART Large} \\ \hline
Softmax (S) & 407.3M & 0.794          & 16.386          & \textbf{0.488}          & 1.091          & 4.654          & 0.359          & 15.386          & 0.476          & 1.006          & 4.136\\
CopyNet~\citep{gu2016incorporating} & 409.4M & 0.774          & 16.268          & 0.485          & 1.113          & 4.619          & 0.358          & 15.293          & 0.473          & 0.995          & 4.144\\
PG~\citep{cnn_dataset_split} & 409.4M & 0.780          & 16.344          & 0.486          & 1.097          & 4.656          & 0.358          & 15.544          & 0.475          & 0.995          & 4.186 \\
PS~\citep{MerityX0S17} & 409.4M & 0.774          & 16.142          & 0.484          & 1.099          & 4.654          & 0.358          & \textbf{15.547} & 0.475          & \textbf{1.000} & 4.227\\
S + CEPR:20 & 414.7M & 0.780          & \textbf{16.394} & \textbf{0.488} & 1.073          & 4.767          & 0.359          & 15.466          & \textbf{0.476} & 0.982          & 4.240 \\
S + CEPR:20 + Mi & 426.2M & \textbf{0.769} & 16.085          & 0.483          & \textbf{1.032} & \textbf{4.811} & \textbf{0.347} & 15.371          & 0.475          & 0.957          & \textbf{4.292}\\ 
\end{tabular}
}
\caption{Comparison of the summaries generated by different models in the test sets of CNN/DM and XSUM datasets. We also report the number of parameters of each model. From top to bottom, the four sections are the results of T5-Small, T5-Base, BART Base, and BART Large. The meaning of the metrics are described in \Cref{sec:gpt-2_more_results}. R2 (ROUGE 2-F1) scores are percentages. Within each section, we highlight the smallest loss, the P Ratio that is closest to 1, and highest numbers in the other metrics.}
\label{tb:summarization_other_news}
\end{table*}

\begin{table*}[t!]
\centering
\scalebox{0.75}{
\begin{tabular}{c|c|ccccc|ccccc|}
&  & \multicolumn{5}{c|}{BookSum Paragraph} & \multicolumn{5}{c|}{SAMSUM} \\
Models & Time (ms) & Loss ($\downarrow$) & R2 & R1P & P Ratio & NIST & Loss ($\downarrow$) & R2 & R1P & P Ratio & NIST \\ \hline
\multicolumn{12}{|c|}{T5-Small} \\ \hline
Softmax (S) & 30.1 & 0.654                         & 1.673                      & 0.149                              & 0.589                           & 1.383                    & 0.383                         & 13.806                     & 0.605                              & 0.873                           & 3.945                    \\
CopyNet~\citep{gu2016incorporating} & 37.0 &0.646                         & 1.722                      & 0.183                              & \textbf{0.747}                  & 1.440                    & 0.381                         & 14.210                     & 0.594                              & 0.809                  & 3.965                    \\
PG~\citep{cnn_dataset_split} & 43.4 & 0.648                         & 1.669                      & 0.160                              & 0.631                           & 1.413                    & 0.392                         & 10.673                     & 0.542                              & 0.711                           & 1.665                    \\
PS~\citep{MerityX0S17} & 37.6 & 0.646                         & 1.627                      & 0.177                              & 0.700                           & 1.417                    & 0.383                         & 13.817                     & 0.583                              & 0.794                           & 3.960                    \\
S + R:20 & 32.9 & 0.652                         & 1.663                      & 0.159                              & 0.677                           & 1.403                    & 0.380                         & 13.728                     & 0.598                              & 0.870                           & 3.995                    \\
S + E & 33.8 & 0.645                         & 1.710                      & 0.171                              & 0.673                           & 1.421                    & 0.370                         & 13.557                     & 0.602                              & 0.892                           & 3.906                    \\
S + CE & 34.0 & 0.644                         & 1.734                      & 0.173                              & 0.680                           & 1.436                    & 0.368                         & 14.136                     & 0.619                              & 0.892                           & 3.971                    \\
S + CER:20 & 35.8 & 0.642                         & 1.710                      & 0.174                              & 0.693                           & 1.434                    & 0.367                         & 14.281                     & 0.627                              & 0.911                           & 3.968                    \\
S + CEPR:20 & 38.4 & \textbf{0.641}                & \textbf{1.768}             & 0.184                     & 0.725                           & \textbf{1.461}           & \textbf{0.365}                & \textbf{14.451}            & \textbf{0.639}                     & 0.909                           & \textbf{4.034}           \\
S + CEPR:20 + Mi & 41.7 & \textbf{0.641}                & 1.733                      & \textbf{0.185}                     & 0.721                  & 1.458                    & \textbf{0.365}                & 14.193                     & 0.630                     & \textbf{0.922}                  & 4.011                   \\ \hline
\multicolumn{12}{|c|}{T5-Base} \\ \hline
Softmax (S) & 102.4 & 0.587          & 1.876          & 0.160          & 0.650          & 1.443          & 0.308          & 17.662          & 0.672          & 0.915          & \textbf{4.559} \\
CopyNet~\citep{gu2016incorporating}  & 110.3 & 0.582          & 1.867          & 0.187          & 0.744          & 1.481          & 0.307          & 17.556          & \textbf{0.678} & 0.901          & 4.544 \\
PG~\citep{cnn_dataset_split} & 117.7 & 0.585          & 1.832          & 0.159          & 0.647          & 1.434          & 0.317          & 14.649          & 0.611          & 0.740          & 1.870 \\
PS~\citep{MerityX0S17} & 112.0 & 0.582          & \textbf{1.899} & 0.176          & 0.718          & 1.465          & 0.308          & 17.502          & 0.660          & 0.897          & 4.453\\
S + CEPR:20 & 115.3 & \textbf{0.580} & 1.842          & \textbf{0.191} & \textbf{0.771} & \textbf{1.482} & \textbf{0.300} & \textbf{18.082} & 0.677          & \textbf{0.950} & 4.553 \\
S + CEPR:20 + Mi & 116.3 & 0.584          & 1.860          & 0.187          & 0.770          & 1.477          & 0.301          & 17.617          & 0.677          & 0.938          & 4.521          \\ \hline
\multicolumn{12}{|c|}{BART Base} \\ \hline
Softmax (S) &  46.6 & 0.624          & 1.807          & 0.141          & 0.656          & 1.425          & 0.327          & \textbf{19.379} & 0.672          & \textbf{0.995} & 4.546 \\
CopyNet~\citep{gu2016incorporating} & 57.8  & \textbf{0.613} & 1.866          & \textbf{0.166} & 0.728          & 1.454          & 0.326          & 18.227          & 0.662          & 0.944          & 4.535 \\
PG~\citep{cnn_dataset_split} & 64.8  &0.624          & 1.864          & 0.140          & 0.668          & 1.428          & 0.328          & 18.791          & 0.673          & 0.963          & 4.537    \\
PS~\citep{MerityX0S17} & 57.9   & \textbf{0.613} & \textbf{1.867} & 0.163          & 0.723          & \textbf{1.461} & 0.324          & 18.367          & 0.674          & 0.951          & 4.573     \\
S + R:20 & 50.5  & 0.627          & 1.807          & 0.154          & 0.720          & 1.430          & 0.326          & 19.022          & 0.671          & 0.971          & \textbf{4.608} \\
S + E &  54.2  &  0.620          & 1.825          & 0.150          & 0.688          & 1.429          & 0.324          & 18.902          & 0.680          & 0.970          & 4.501     \\
S + CE & 56.5 & 0.619          & 1.847          & 0.153          & 0.685          & 1.441          & 0.323          & 18.739          & 0.672          & 0.949          & 4.537     \\
S + CER:20 & 57.2 & 0.618          & 1.834          & 0.156          & 0.727          & 1.444          & \textbf{0.321} & 19.267          & \textbf{0.678} & 0.981          & 4.561      \\
S + CEPR:20 & 58.8 &  0.618          & 1.865          & 0.157          & \textbf{0.742} & 1.457          & \textbf{0.321} & 18.631          & 0.670          & 0.992          & 4.516     \\
S + CEPR:20 + Mi & 63.2 & 0.620          & 1.827          & 0.158          & 0.733          & 1.442          & 0.322          & 18.681          & 0.670          & 0.987          & 4.439   \\ \hline
\multicolumn{12}{|c|}{BART Large} \\ \hline
Softmax (S) & 143.5 & 0.554          & \textbf{2.094} & 0.171          & 0.722          & 1.472          & \textbf{0.303} & 20.848 & 0.711 & \textbf{1.006} & 4.621 \\
CopyNet~\citep{gu2016incorporating} & 168.9 & 0.548 & 2.087          & \textbf{0.184}          & 0.762          & 1.490          & 0.298 & 21.703 & 0.708 & 1.026 & 4.727\\
PG~\citep{cnn_dataset_split} &  178.3 & 0.731          & 2.090          & 0.174          & 0.725          & 1.479          & 0.301 & 21.428 & 0.706 & 1.051 & 4.604\\
PS~\citep{MerityX0S17} & 168.5 & 0.726          & 2.083          & \textbf{0.184} & 0.760          & 1.493          & 0.300 & \textbf{22.144} & 0.710 & 1.036 & \textbf{4.779}\\
S + CEPR:20 & 169.9 & 0.552          & 2.069          & 0.178          & \textbf{0.763} & \textbf{1.505} & 0.302 & 21.326 & 0.691 & 1.017 & 4.595  \\
S + CEPR:20 + Mi & 177.4 & \textbf{0.544}          & 2.024          & 0.175          & 0.737          & 1.500          & 0.294 & 21.244 & \textbf{0.713} & 0.959 & 4.746 \\
\end{tabular}
}
\caption{Comparison of the summaries generated by different models in the test sets of BookSum and SAMSUM datasets. We also report the inference time of one samples. The meaning of the metrics are described in \Cref{sec:gpt-2_more_results}. R2 (ROUGE 2-F1) scores are percentages. Within each section, we highlight the smallest loss, the P Ratio that is closest to 1, and highest numbers in the other metrics.}
\label{tb:summarization_other_types}
\end{table*}

\subsection{Summarization}

Compared to \autoref{fig:gpt_loss_size}, \autoref{fig:t5_loss_size} shows that our methods improve the loss of T5 in CNN/DM more than GPT-2 in Wikipedia.

In \autoref{tb:summarization_other_news} and \autoref{tb:summarization_other_types}, we compare the different summarization models by their model size,  evaluation losses, inference time, and other metrics which we use in \autoref{sec:gpt-2_more_results}. The pointer network baselines and our methods significantly improve most metrics over the softmax baseline, which is used ubiquitously in nearly all LMs. Although our method generally improves less on the T5-Base model, the percentages of additional parameters and inference time overhead are much smaller. Although our methods tend to improve less in larger language model, we still improve BART Large very significantly in NIST, CIDEr, and MAUVE, and Mi seems to become more effective in BART Large. 

The testing set of SAMSUM dataset only has 819 samples, so some metrics such as R1 and R2 are not as stable as other three datasets.
PG~\citep{cnn_dataset_split} for T5-Small and T5-Base perform much worse in SAMSUM dataset. We hypothesize that it is because the dialog input in SAMSUM dataset is very different from the pretraining data of T5, which makes training PG unstable.

In most datasets and models, the R Ratio from our method is significantly closer to 1 than the softmax baseline, which means the average number of proper nouns in our summaries is much closer to the average number of proper nouns in the human-written summary. For example, in BookSum Paragraph, we improve its R Ratio by 26\%, which partially explains our large MAUVE improvement in \autoref{tb:summarization}. Notice that our methods do not always output more proper nouns. For example, for BART Base in CNN/DM dataset, our methods reduce the R Ratio of the softmax baseline, which is larger than 1. This shows that our methods could learn when we should copy the proper nouns according to the training data.

%% file: appendix/method_details.tex
\section{Method Details}
\label{sec:method_details}
We describe some details of our methods and baselines in this section.

\subsection{Proposed Methods}

To allow us to start from existing LMs that are pretrained using softmax, we keep the modified softmax layer initially working almost the same as the original softmax layer. We initialize the linear transformation weights of $L_{PD}^f()$, $L_{LD}^f()$, $L_{PE}^f()$, and $L_{LE}^f()$ as $10^{-10} \cdot \mathbb{I}$. The other linear weights $L_{.}^f()$ are initialized as the identity matrix $\mathbb{I}$.

In the local decoder embedding method \textbf{Softmax + P + Mi}, the initialization would give the 0 logit to all context words. To solve the issue, we revise \autoref{eq:logit_general} a little and compute $\text{Logit}_{P}(x,c_t)$ by
\begin{equation}
\label{eq:logit_P}
\left\{
\begin{matrix*}[l]
\vf_{c_t,V}^T \vw_x + \vf_{c_t,PD}^T \vf_{x,c_t,LD} \;\; \text{if} \; x \in c_t\\ 
\vf_{c_t,V}^T \vw_x \;\; \text{O/W}
\end{matrix*}\right..
\end{equation}
That is, we initially rely on the original softmax layer to compute all the logits and let the term $\vf_{c_t,PD}^T \vf_{x,c_t,LD}$ gradually influences the logits of the context words.

In \textbf{MoS + CPR:20,100 + Mi}, our proposed method only revises the logit in one of the softmax.




\subsection{Pointer Network Baselines}
\label{sec:pointer_imp}


The pointer networks are originally designed for RNN, so we are unable to use exactly the same formula proposed in the papers. Nevertheless, we try our best to adapt the pointer networks for the transformer encoder while keeping the gist of the formulas. In all methods, to let the results more comparable to our methods, we use $\vf_{c_t,PE}$ and $L_{LE}^f$ to determine the probability of copying the words from the context, and use $\vf_{c_t,V}^T \vw_x$ to determine the probability of generating all the words in the vocabulary.


In CopyNet~\citep{gu2016incorporating}, we compute the probability of outputting the word $x$ as
\begin{align}
\label{eq:CopyNet}
Prob(x | I, c_t) \propto \text{exp}\left(\vf_{c_t,V}^T \vw_x\right) \nonumber \\ 
+ \sum_{j=1}^{|I|} \mathbbm{1}_{ {I^j}=x} \text{exp}\left(\vf_{c_t,PE}^T L_{LE}^f(\vh_{I^j}^M) + b\right).
\end{align}

Notice that CopyNet needs to sum up the exponential of dot products, which often causes overflow problems in GPT-2. We can set $b$ to be a large negative value initially to solve the problem, but its perplexity is much worse than the other two pointer network variants. Thus, we choose to skip the CopyNet in the GPT-2 experiments.

In Pointer Generator~\citep{cnn_dataset_split}, we compute the probability of $x$ using
\begin{align}
\label{eq:Pointer}
Prob(x | I, c_t) = p_{gen} \frac{ \text{exp}\left(\vf_{c_t,V}^T \vw_x\right)}{Z_V} \nonumber \\ 
+ (1-p_{gen}) \sum_{j=1}^{|I|} \mathbbm{1}_{ {I^j}=x} P_E(j | I, c_t) ,
\end{align}
where $P_E(j | I, c_t) = \frac{ \text{exp}\left( \vv^T \text{tanh}(\vf_{c_t,PE} + L_{LE}^f(\vh_{I^j}^M) + \vb) \right)}{Z_E}$, $p_{gen} = \sigma(\vq^T \vh_{c_t}^M + b_{ptr})$, the normalization term $Z_V = \sum_{x \in V} \text{exp}\left(\vf_{c_t,V}^T \vw_x\right)$, and $Z_E = \sum\limits_{j=1}^{|I|} \text{exp}\left( \vv^T \text{tanh}(\vf_{c_t,PE} + L_{LE}^f(\vh_{I^j}^M) + \vb) \right)$.

We skip the coverage mechanism in the pointer generator paper to make it more comparable to other methods. In T5 experiments, its training loss is sometimes very large, so we set $b_{ptr}$ as 3 initially to keep the $p_{gen}$ close to 1 (i.e., turn pointer part off initially). In other experiments, we set $b_{ptr}=0$.

\begin{figure*}[t!]
\centering
\includegraphics[width=0.9\linewidth]{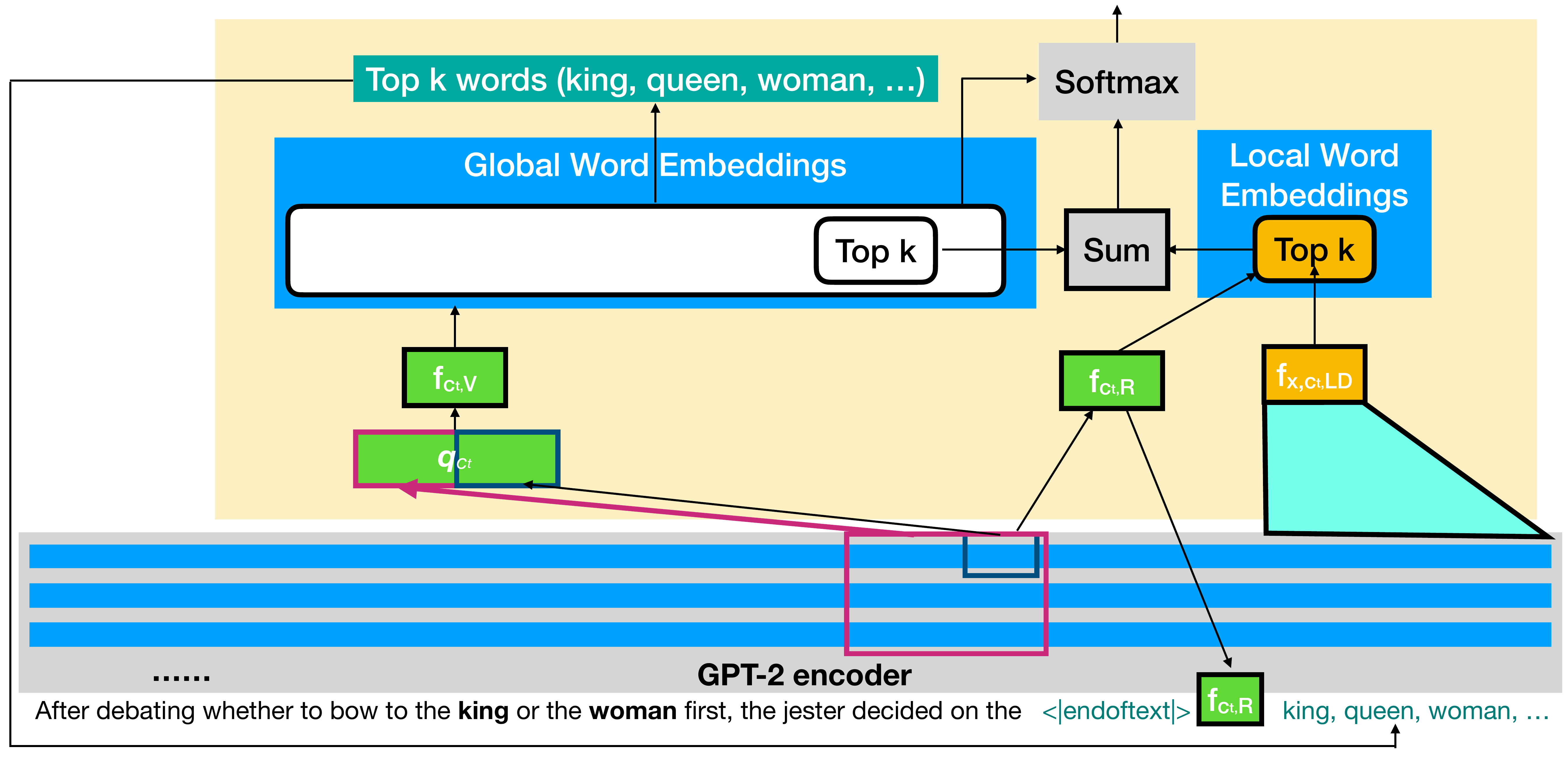}
\caption{Word-by-word reranker architecture. 
}
\label{fig:reranker_stage2}
\end{figure*}

\begin{figure}[t]
\centering
\includegraphics[width=0.9\linewidth]{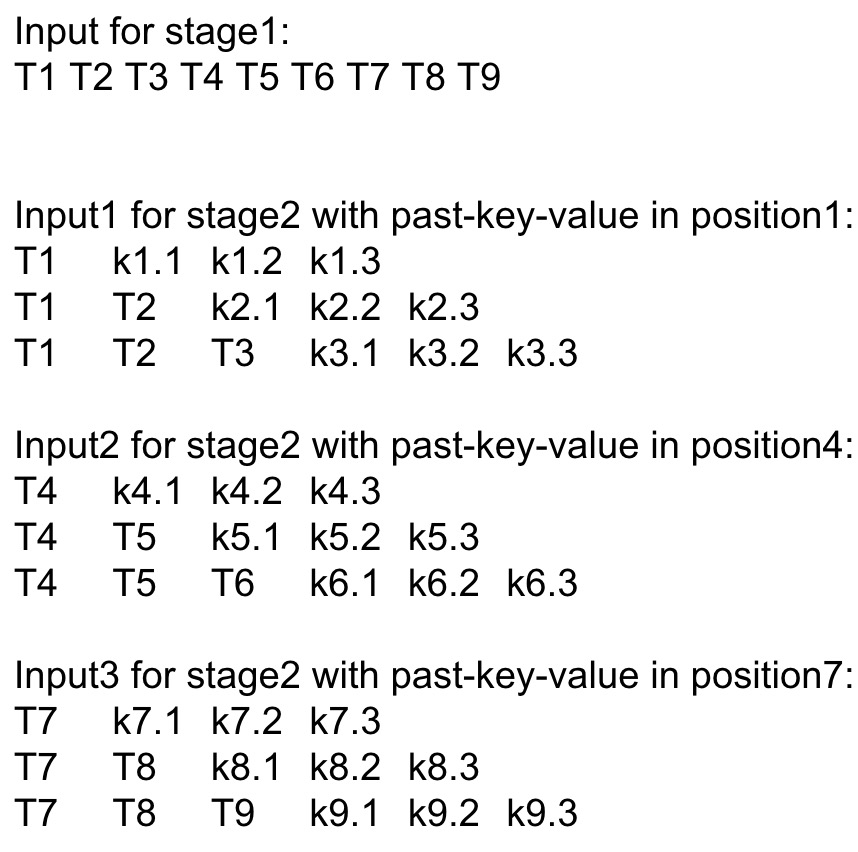}
\caption{Our efficient implementation of word-by-word reranker. Ti is tokens and ki.j is top-k tokens for Ti.}
\label{fig:reranker_stage2_input}
\end{figure}

In Pointer Sentinel~\citep{MerityX0S17}, the probability of $x$ is computed by
\small
\begin{align}
\label{eq:Pointer_sentinel}
Prob(x | I, c_t) = g \frac{ \text{exp}\left(\vf_{c_t,V}^T \vw_x\right)}{Z_V} \nonumber \\ 
 + \sum_{j=1}^{|I|} \mathbbm{1}_{ {I^j}=x} \frac{ \text{exp}\left(\vf_{c_t,PE}^T \text{tanh}(L_{LE}^f(\vh_{I^j}^M)) + b\right)}{Z_p} ,
\end{align}
\normalsize
$g = \frac{ \text{exp}(\vq^T \vh_{c_t}^M) }{Z_p}$, and $Z_p = \text{exp}(\vq^T \vh_{c_t}^M) + \sum_{j=1}^{|I|} \text{exp}\left(\vf_{c_t,PE}^T \text{tanh}(L_{LE}^f(\vh_{I^j}^M)) + b\right)$.

In our experiments, we find that the pointer network variants usually have similar performance (except that PG sometimes performs much worse in summarization due to some training stability issues). This suggests that the differences in the pointer network variants often do not influence the performance significantly, which justifies our simplification of the formulas in the original paper and supports our conclusion that the improvement comes from breaking the softmax bottleneck. 

Notice that in the above pointer network variants, the pointer part can only increase the probability of the context words from the generator part. As a result, it cannot alleviate the repetition problem in the last example of \autoref{tb:context_example}.

\subsection{Word-by-word Reranker Baseline}
\label{sec:wbw_rerank}

We illustrate our word-by-word reranker (wbwR) in \autoref{fig:reranker_stage2}. The method has two stages. In the first stage, we compute the logits using the projected hidden state $\vf_{c_t,V}$ and retrieve the top $k$ words. At the second stage, we append the top $k$ words to the input context along with the hidden state $\vf_{c_t,R}$ for reranking the context words.\footnote{The motivation is helping GPT-2 to output the local word embedding of a candidate closer to the $\vf_{c_t,R}$ if GPT-2 wants to increase the probability of the candidate.} We use the same positional embeddings for all candidates to encourage the model to change the ranking of the words. Next, we use the hidden states corresponding to the candidates to compute their local word embeddings as $\vf_{x,c_t,LD}$. Finally, we re-estimate the probabilities of top $k$ words by 
\begin{equation}
\label{eq:logit_w2wR}
\left\{
\begin{matrix*}[l]
\vf_{c_t,V}^T \vw_x + \vf_{c_t,R}^T \vf_{x,c_t,LD} \;\; \text{if} \; x \in W(k)\\ 
\vf_{c_t,V}^T \vw_x \;\; \text{O/W}
\end{matrix*}\right..
\end{equation}
To improve the quality of our top k candidates, the final loss is the addition of the wbwR loss at the second stage and the loss of the original softmax layer that only uses the logits from $\vf_{c_t,V}^T \vw_x$ at the first stage. When we combine the wbwR with \textbf{Softmax + CPR:20,100 + Mi}, we simply use \textbf{Softmax + CPR:20,100 + Mi} at the first stage and use the wbwR to overwrite the logits of \textbf{Softmax + CPR:20,100 + Mi} at the second stage.

Using this method, we can update the embeddings of the words that are not in the context and allow the candidates to interact with the input context to determine their probabilities as the classic two-stage reranker while keeping the model size roughly unchanged. Nevertheless, the method can only change the probability of the top $k$ words and its computational overhead and memory requirement prevents us from using a very large $k$.


Unlike the standard GPT-2, we cannot get the probability of all positions in one forward pass because the input contexts are different when computing the probability at each position and the input of the second stage reranker depends on the results of the previous forward pass at the first stage. To speed up, we reuse the computed hidden states and batchify the forward passes.

In our implementation, we first get the top $k$ candidates corresponding to all tokens in the stage1 (just original GPT2) as the input of stage2 reranker. To avoid recalculating the hidden states of the context at stage2, we store the hidden states using the past-key-value in Hugging Face and only compute the hidden states corresponding to the top $k$ candidate tokens at stage2. 

We divide the computation of the whole input sequence into several blocks as shown in Figure \ref{fig:reranker_stage2_input}. In each block, we  input a batch containing the last few tokens and top $k$ candidates into the GPT-2 while reusing the hidden states of their common contexts from stage1. In this way, we can increase parallelism by increasing the block size if the GPU memory allows it.

Even though we spent substantial effort on optimizing the wbwR, the method is still too slow to be practically useful. Even if we use four RTX 8000 (a faster GPU with a larger memory), our wbwR implementation is still around 10 times slower than our proposed \textbf{Softmax + CPR:20,100 + Mi} that uses only one RTX 2080.

%% file: appendix/exp_details.tex
\section{Experiment Details}
\label{sec:exp_details}
For the reproducibility, we provide some experimental configuration in this section. Please see our codes (\url{https://github.com/iesl/Softmax-CPR}) for more details.

\subsection{GPT-2 Experimental Details}
We mostly follow the experimental setup \citet{chang2022softmax} except that we share the input and output word embeddings as in the standard GPT-2 models. As in \citet{chang2022softmax}, we use the last 2\% of the corpus as the test set and the 2\% before that as the validation set.\footnote{We do not shuffle the corpus before splitting the datasets. We found that our improvement could be even larger if we shuffle the corpus to let the training data distribution closer to the testing data distribution.} In the word-by-word reranking experiment, we only use first 100k tokens in validation set to speed up the evaluation. To show that our model could be added to existing pretrained models, we continue training the pretrained GPT-2. For GPT-2 Small, we train for $1$ epoch, and for GPT-2 Medium, we train for $0.4$ epoch. We find that the performance improvements usually do not change significantly after training for $0.1$ epoch. 

As in \citet{chang2022softmax}, we set the sequence length as $200$, batch size as $4$, and learning rate for AdamW as $1e-5$. Our methods only have two hyperparameters, $k_1$ and $k_2$, and we try values $20$, $100$, $200$, $500$ and select $20$ and $100$ using the validation data.





In the text completion experiment, we generate $360$k continuations with a length of $50$ given the prompts in Wikipedia. We first sample $40$k sequences in the test data of Wikipedia 2021. Next, we use the first $20$, $70$, and $120$ words in the sequence as our context and let the different models generate the next $50$ words as continuations. The references are the actual next $50$ words. All the methods use Top-K sampling and K=$5$.


\begin{table}[t]
    \centering
    \begin{tabular}{l|ccc}
    
    & \footnotesize{Train} & \footnotesize{Val} & \footnotesize{Test}
    \\ \hline
    \footnotesize{CNN/DM} & \footnotesize{287113} & \footnotesize{13368} & \footnotesize{11490}
    \\ 
    \footnotesize{XSUM} & \footnotesize{204045} & \footnotesize{11332} & \footnotesize{11334}
    \\ 
    \footnotesize{BookSum Paragraph} & \footnotesize{210931} & \footnotesize{27222} & \footnotesize{26025}
    \\ 
    \footnotesize{SAMSUM} & \footnotesize{14732} & \footnotesize{818} & \footnotesize{819}
    \\
    \end{tabular}
    \caption{Dataset size of our four summarization tasks.} 
    \label{summarization-train-valid-test}
\end{table}

\subsection{Summarization Experimental Details}

BookSum dataset~\citep{kryscinski2021booksum} includes three summarization tasks: Summarizing a book, a chapter, and a paragraph. We test our methods using the paragraph summarization task due to the input length restriction of BART and T5. The dataset is constructed by 
automatically aligning the paragraphs in a chapter with the sentences in a chapter summary, which introduces noise to the dataset. Similarly, XSUM uses the first sentence in news instead of manually-written summary as the ground truth reference.
The relatively noisier datasets such as XSUM and BookSum Paragraph, and smaller dataset like SAMSUM could test the stability of the methods. The sizes of the summarization datasets could be found in
\autoref{summarization-train-valid-test}.


We conduct the summarization experiments based on a summarization example code from Hugging Face\footnote{\url{https://github.com/huggingface/transformers/blob/main/examples/pytorch/summarization/run_summarization.py}}. Most of our hyperparameters use the default value in the code. In our preliminary study, our improvement is not sensitive to the hyperparameter choice (e.g., the improvement gap is similar across different numbers of epochs). Thus, we do not tune the hyperparameters for each method or for each dataset unless we cannot reach a low training loss at the end.

In CNN/DM, XSUM, and SAMSUM datasets, We train models for $3$ epochs. In BookSum datasets, We trained models for $5$ epochs.\footnote{The BookSum Paragraph is noisier, so we train longer to be safe. And we find that different numbers of epochs do not change the trend of the results.} The learning rate is set to be $5e-05$ except for BART Large model in BookSum, where we use $1e-05$ to stablize the training of all methods. 

All the experiments use batch size 8 and AdamW with betas=($0.9$,$0.999$), epsilon=$1e-06$, weight-decay=$1.2e-6$. During the generation, we used Top-K sampling (K=$10$) as our decoding method. The maximum summary length is set as $128$ and maximum input length is $1024$. We use warmup for the first $1000$ steps in all the experiments, which allows us to change the architecture of T5 and BART more significantly (e.g., using Mi) without having a training stability issue.



The $k$ in the reranker partition and the block size of multiple input hidden states (Mi) is coarsely tuned based on validation performance of CNN/DM. Unlike considering the top $100$ words in the open-end text completion using GPT-2, we find that reranking the top $20$ words is sufficient for our summarization models, probably because next words are easier to predict in the summarization task.

For our evaluation metrics, we use the default setting for ROUGE\footnote{\url{https://huggingface.co/spaces/evaluate-metric/rouge}} and set use\_stemmer=True. When reporting the ROUGE scores, we follow the conventions to show their percentages. We use the default setting for
MAUVE\footnote{\url{https://huggingface.co/spaces/evaluate-metric/mauve}}, CIDER\footnote{\url{https://github.com/vrama91/cider}}, NIST\footnote{\url{https://www.nltk.org/api/nltk}}. For MAUVE, we insert a new line symbol after every sentence as in the original Hugging Face summarization example code.
For factCC metric\footnote{\url{https://github.com/salesforce/factCC}}, we use the author-provided checkpoint to evaluate CNN/DM results since factCC is originally trained in CNN/DM. For the other three summarization datasets, we follow the author's codes to constructed positive and negative data and continued training the CNN/DM factCC model on each dataset with one epoch respectively. Then we evaluate different summarization tasks with the corresponding factCC checkpoint.

\subsection{Computational Environment and Software}
We implement our methods by revising the Hugging Face library~\cite{Wolf_Transformers_State-of-the-Art_Natural_2020}. From Hugging Face, we load the pretrained LMs including
GPT-2 Small\footnote{\url{https://huggingface.co/gpt2}}, GPT-2 Medium\footnote{\url{https://huggingface.co/gpt2-medium}}, T5-Small\footnote{\url{https://huggingface.co/t5-small}}, T5-Base\footnote{\url{https://huggingface.co/t5-base}}, BART Base\footnote{\url{https://huggingface.co/facebook/bart-base}}, and BART Large\footnote{\url{https://huggingface.co/facebook/bart-large}}. We use SpaCy~\citep{Honnibal_spaCy_Industrial-strength_Natural_2020} to detect the proper nouns.

For GPT-2 Medium, T5-Base, and BART Large, we use NVIDIA GeForce RTX 8000 to train the model and for other smaller models, we use NVIDIA GeForce RTX 2080. Most of experiments could be done within one week. In all the inference time experiments, we use NVIDIA GeForce GTX TITAN X, batch size 4 for GPT-2, and batch size 8 for BART and T5.


